\documentclass{article}
\pdfoutput=1

\usepackage{algorithm}
\usepackage{algpseudocode}
\usepackage{amsmath}
\usepackage{amssymb}
\usepackage{booktabs}
\usepackage{cleveref}
\usepackage{enumitem}
\usepackage{graphicx}
\usepackage{multirow}
\usepackage{natbib}
\usepackage[main, final]{neurips_2025}
\usepackage{pdfpages}
\usepackage{subfigure}
\usepackage{url}
\usepackage{xcolor}

\newtheorem{assumption}{Assumption}

\newtheorem{remark}{Remark}

\title{ESCORT: Efficient Stein-variational and Sliced Consistency-Optimized Temporal Belief Representation for POMDPs}

\author{%
    Yunuo Zhang \\
    Vanderbilt University \\
    yunuo.zhang@vanderbilt.edu\\
    \And
    Baiting Luo \\
    Vanderbilt University \\
    baiting.luo@vanderbilt.edu\\
    \And
    Ayan Mukhopadhyay \\
    Vanderbilt University \\
    ayan.mukhopadhyay@vanderbilt.edu\\
    \AND
    Gabor Karsai \\
    Vanderbilt University \\
    gabor.karsai@vanderbilt.edu\\
    \And
    Abhishek Dubey \\
    Vanderbilt University \\
    abhishek.dubey@vanderbilt.edu\\
}

\begin{document}
\maketitle

\begin{abstract}
In Partially Observable Markov Decision Processes (POMDPs), maintaining and updating belief distributions over possible underlying states provides a principled way to summarize action-observation history for effective decision-making under uncertainty. As environments grow more realistic, belief distributions develop complexity that standard mathematical models cannot accurately capture, creating a fundamental challenge in maintaining representational accuracy. Despite advances in deep learning and probabilistic modeling, existing POMDP belief approximation methods fail to accurately represent complex uncertainty structures such as high-dimensional, multi-modal belief distributions, resulting in estimation errors that lead to suboptimal agent behaviors. To address this challenge, we present ESCORT (Efficient Stein-variational and sliced Consistency-Optimized Representation for Temporal beliefs), a particle-based framework for capturing complex, multi-modal distributions in high-dimensional belief spaces. ESCORT extends SVGD with two key innovations: correlation-aware projections that model dependencies between state dimensions, and temporal consistency constraints that stabilize updates while preserving correlation structures. This approach retains SVGD's attractive-repulsive particle dynamics while enabling accurate modeling of intricate correlation patterns. Unlike particle filters prone to degeneracy or parametric methods with fixed representational capacity, ESCORT dynamically adapts to belief landscape complexity without resampling or restrictive distributional assumptions. We demonstrate ESCORT's effectiveness through extensive evaluations on both POMDP domains and synthetic multi-modal distributions of varying dimensionality, where it consistently outperforms state-of-the-art methods in terms of belief approximation accuracy and downstream decision quality. Our code is available at https://github.com/scope-lab-vu/ESCORT.
\end{abstract}

\section{Introduction} \label{sec:introduction}
    Partially Observable Markov Decision Processes (POMDPs)~\citep{pomdps_definition_aström} provide a powerful mathematical framework for sequential decision-making under uncertainty, enabling agents to make optimal decisions despite having only partial information about their environment~\citep{pomdps_framework_kaelbling}. At the core of POMDP solutions lies the concept of belief states: probability distributions over possible underlying states conditioned on the history of actions and observations~\citep{pomdps_framework_kaelbling}. As POMDPs are applied to increasingly complex real-world domains~\citep{pomdps_robotics_lauri,pomdps_robotics_kurniawati,shrinking_pomcp_zhang}, the underlying belief distributions develop sophisticated characteristics that standard mathematical models struggle to accurately capture~\citep{pomdps_pca_roy,parametric_belief_brooks, airoas_zhang}. Specifically, realistic belief distributions in POMDPs often exhibit a challenging combination of high dimensionality (due to complex state spaces)~\citep{pomdps_pca_roy} and multi-modality (from ambiguous observations creating multiple distinct hypotheses)~\citep{forbes_chen, airoas_zhang}, which traditional approaches struggle to model efficiently. The inability to efficiently represent and update these complex belief distributions creates a fundamental bottleneck in developing practical POMDP solutions, as even small errors in belief approximation can propagate and significantly degrade decision quality over time.

    Existing approaches to belief approximation in POMDPs face significant limitations with complex distributions. Parametric methods using neural representations struggle with uncertainty structures: DRQN~\citep{drqn_hausknecht} and ADRQN~\citep{adrqn_zhu} compress histories into vectors that poorly capture multi-modal uncertainty, while even DVRL~\citep{dvrl_igl}, despite using particle-based VAEs~\citep{vae_kingma}, fails to maintain multiple distinct hypotheses simultaneously. These parametric approaches efficiently process high-dimensional data but sacrifice representational expressiveness---despite theoretical universal approximation power, neural networks face computational inefficiency and generalization challenges~\citep{deep_learning_generalization_zhang}. Their fixed parametric nature prevents adaptation to varying uncertainty complexity, causing cumulative belief estimation errors over time.

    On the other hand, particle-based methods offer flexibility in representing arbitrary distributions but face critical limitations. SIR filters~\citep{particle_filters_gordon}, which underpin leading POMDP solvers like POMCP~\citep{pomcp_silver}, POMCPOW~\citep{pomcpow_sunberg}, ARDESPOT~\citep{ardespot_somani}, and AdaOPS~\citep{adaops_wu}, struggle with the curse of dimensionality and particle degeneracy in high-dimensional spaces. Their stochastic resampling leads to mode collapse, failing to maintain coverage across multi-modal distributions, especially with ambiguous observations~\citep{airoas_zhang}. These methods also inefficiently capture dependencies between state dimensions—either making oversimplified independence assumptions or requiring exponentially more particles to model joint distributions accurately, significantly limiting their applicability to complex POMDPs.

    Inspired by the effectiveness of Stein Variational Gradient Descent (SVGD)~\citep{svgd_liu} in Bayesian inference, we explore deterministic particle evolution as a principled alternative. SVGD avoids resampling-induced degeneracy through continuous gradient-based updates: particles move deterministically via $\nabla\log p(x)$ with kernel repulsion $\nabla k(x,x')$ maintaining multi-modal coverage without discarding hypotheses~\citep{svgd_liu}. Unlike fixed parametric architectures, SVGD dynamically adapts its particle distribution—concentrating particles in high-uncertainty regions while providing sparse coverage elsewhere—aligning representational capacity with belief complexity without architectural changes. However, standard SVGD itself suffers from kernel degeneracy in high-dimensional spaces~\citep{mpsvgd_zhuo,psvgd_chen}—weakening both attractive and repulsive forces—and cannot preserve complex correlation structures between state dimensions, leading to mode collapse in multi-modal distributions. Recent extensions like MP-SVGD~\citep{mpsvgd_zhuo} and SVMP~\citep{svmp_wang} have demonstrated success in high-dimensional Bayesian inference by leveraging graphical model structures to guide particle evolution. However, these methods require fixed structures that cannot adapt to observation-dependent correlations in POMDPs, where belief correlation patterns change dynamically with observation history~\citep{dimensions_correlation_boyen}.

    ESCORT addresses these fundamental limitations through a novel belief update mechanism that extends SVGD with two key regularization components. Drawing insights from sliced optimal transport theory~\citep{gswd_kolouri}, we introduce: (1) a correlation-aware regularization that preserves dimensional dependencies during particle updates, mitigating kernel degeneracy while maintaining multi-modal representational flexibility, and (2) a temporal consistency constraint that prevents unrealistic belief jumps between timesteps while preserving learned correlation structures. This deterministic update framework with targeted regularization prevents the accumulation of estimation errors that propagate through sequential decision-making. Our contributions are:
    \begin{itemize}[leftmargin=*,itemsep=2pt,topsep=2pt,parsep=0pt]
        \item We extend SVGD to overcome particle degeneracy in traditional filters and fixed representational capacity in parametric approaches for complex POMDP belief approximation.
        \item We introduce correlation-aware regularization inspired by optimal transport theory that preserves dimensional dependencies and mitigates kernel degeneracy in high-dimensional spaces.
        \item We develop temporal consistency regularization that prevents unrealistic belief jumps while preserving correlation structures, ensuring stable belief evolution.
        \item While future work will address computational overhead from correlation matrix computation and projection optimization, ESCORT provides a modular belief representation that seamlessly integrates with existing POMDP solvers for broader practical impact.
        \item We demonstrate ESCORT's effectiveness through extensive experiments on Light-Dark Navigation~\citep{light_dark_platt,pomcp_silver}, Kidnapped Robot~\citep{kidnapped_choset}, and Multi-Target Tracking~\citep{target_tracking_li} benchmarks, as well as synthetic multi-modal distributions, showing consistent improvements in belief fidelity and decision quality.
    \end{itemize}

\section{Background} \label{sec:background}
    \subsection{Partially Observable Markov Decision Processes (POMDPs)} \label{sec:pomdps}
        A partially observable Markov decision process (POMDP)~\citep{pomdps_definition_aström} is formalized as a tuple $\langle S, A, T, R, \Omega, O, \gamma \rangle$: states $S$, actions $A$, transition function $T(s'|s,a)$ (probability of transitioning to $s'$ from state $s$ via action $a$), reward function $R(s,a)$, observations $\Omega$, observation function $O(o|s',a)$ (probability of observing $o$ after reaching $s'$ via action $a$), and discount factor $\gamma \in [0,1)$. Unlike MDPs, agents cannot directly observe states, fundamentally increasing problem complexity.

        Agents maintain belief states $b(s)$ - probability distributions over possible states. After action $a$ and observation $o$, beliefs update via Bayes' rule: $b'(s') = \eta \cdot O(o|s',a) \sum_{s \in S} T(s'|s,a)b(s)$, where $\eta$ normalizes to ensure $\sum_{s' \in S} b'(s') = 1$. This update encapsulates the agent's knowledge given their action-observation history.

    \subsection{Stein Variational Gradient Descent (SVGD)} \label{sec:svgd}
        SVGD~\citep{svgd_liu} is a deterministic sampling algorithm that iteratively transports particles $\{x_i\}_{i=1}^n$ to approximate a target distribution $p(x)$ by minimizing KL divergence through functional gradient descent. Particles update via $x_i \leftarrow x_i + \epsilon \phi(x_i)$, where $\epsilon$ is step size and the optimal velocity field $\phi^*(x) = \frac{1}{n}\sum_{j=1}^n [k(x_j,x)\nabla_{x_j} \log p(x_j) + \nabla_{x_j}k(x_j,x)]$ belongs to a reproducing kernel Hilbert space with kernel $k(x,x')$~\citep{svgd_liu}.

        The update balances exploitation ($k(x_j,x)\nabla_{x_j} \log p(x_j)$ driving particles toward high-density regions) and exploration ($\nabla_{x_j}k(x_j,x)$ creating repulsive forces preventing collapse). The RBF kernel $k(x,x') = \exp(-\frac{1}{h}||x-x'||^2)$ with bandwidth $h$ controls inter-particle interactions~\citep{rbf_median_garreau,svgd_liu}, offering theoretical convergence guarantees with computational efficiency.
        
        However, SVGD struggles in high-dimensional POMDPs where correlation structures $C_{ij} = \frac{\Sigma_{ij}}{\sqrt{\Sigma_{ii}\Sigma_{jj}}}$ (with $i,j$ indexing state dimensions) capture statistical dependencies~\citep{forbes_chen}. Standard isotropic RBF kernels create uniform repulsive forces~\citep{mpsvgd_zhuo}, failing to preserve anisotropic belief distributions' correlation patterns. This causes kernel degeneracy in high dimensions~\citep{psvgd_chen}—kernel values become nearly constant—preventing particles from aligning along principal correlation directions, degrading belief approximation and POMDP performance.

    \subsection{Projection Methods in Optimal Transport} \label{sec:gswd}
        Optimal Transport (OT) provides a geometric framework for comparing probability distributions via minimal transformation cost~\citep{optimal_transport_villani}. The Wasserstein distance $W_p(\mu, \nu) = \left(\inf_{\gamma \in \Gamma(\mu, \nu)} \int \|x-y\|^p d\gamma(x,y)\right)^{1/p}$ captures both probability mass differences and geometric relationships, where $\Gamma(\mu, \nu)$ denotes joint distributions with marginals $\mu$ and $\nu$.

        To reduce computational expense in high dimensions, Sliced Wasserstein Distance (SWD)~\citep{sliced_wasserstein_rabin} projects distributions onto one-dimensional subspaces: $\text{SW}_p(\mu, \nu) = \left(\int_{\mathbb{S}^{d-1}} W_p^p(\mathcal{R}[\mu]_\theta, \mathcal{R}[\nu]_\theta) d\sigma(\theta)\right)^{1/p}$, where $\mathcal{R}[\cdot]_\theta$ denotes the Radon transform along direction $\theta$. Generalized Sliced Wasserstein (GSW) Distance~\citep{gswd_kolouri} extends this with nonlinear transformations: $\text{GSW}_p(\mu, \nu) = \left(\int_{\Theta} W_p^p(\mathcal{R}_h[\mu]_\theta, \mathcal{R}_h[\nu]_\theta) d\lambda(\theta)\right)^{1/p}$ using parameterized function $h_\theta(x)$, with max-GSW Distance variant: $\text{max-GSW}_p(\mu, \nu) = \max_{\theta \in \Theta} W_p(\mathcal{R}_h[\mu]_\theta, \mathcal{R}_h[\nu]_\theta)$.
        
        ESCORT integrates these projection principles as regularization mechanisms within SVGD's update process rather than as distance metrics. This enables learning transformation matrices that identify significant correlation directions while mitigating kernel degeneracy, allowing particles to align along principal correlation directions while maintaining multi-modal diversity—addressing fundamental challenges in complex POMDP belief representation.

\section{Approach} \label{sec:approach}
    \begin{figure}[htbp]
        \centering
        \includegraphics[width=1.0\textwidth]{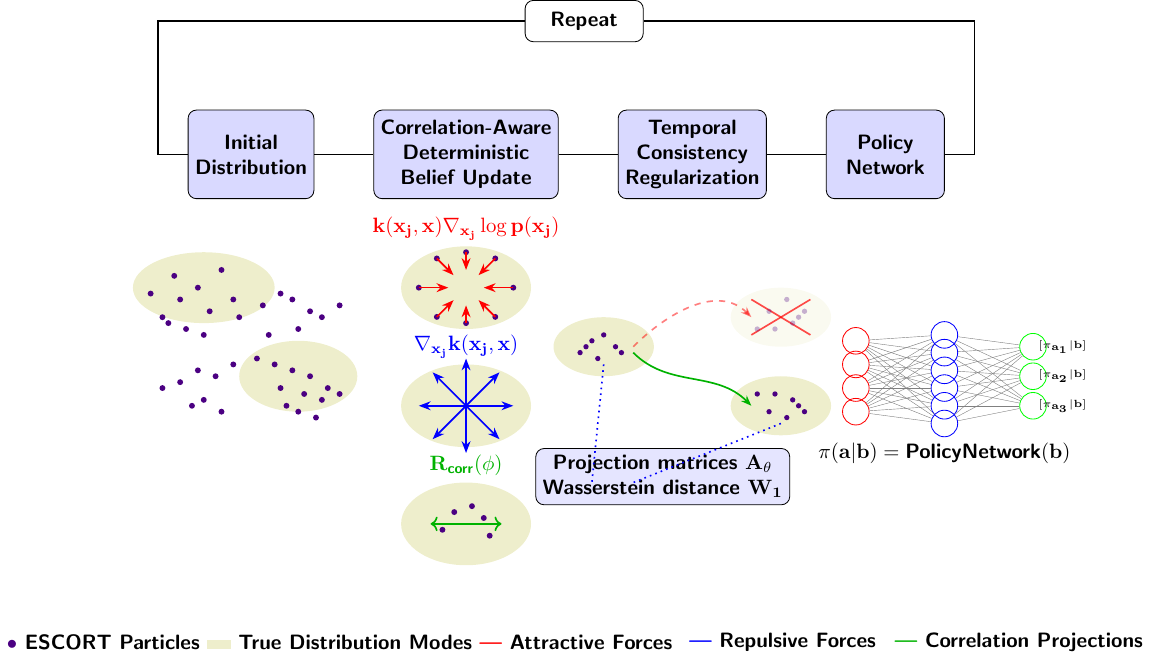}
        \caption{Overview of the ESCORT framework. The diagram illustrates the iterative process of maintaining accurate belief representations in POMDPs through deterministic particle evolution. Purple dots represent particles, yellow regions show distribution modes, and colored arrows indicate different force types. In the Temporal Consistency component, green arrows represent permissible belief transitions while red crossed arrows indicate prevented unrealistic jumps.}
        \label{fig:framework_overview}
    \end{figure}

    We present ESCORT (\textbf{E}fficient \textbf{S}tein-variational and sliced \textbf{C}onsistency-\textbf{O}ptimized \textbf{R}epresentation for \textbf{T}emporal beliefs), a particle-based framework addressing the challenges of belief approximation in complex POMDPs, as illustrated in Figure~\ref{fig:framework_overview}. ESCORT extends SVGD with correlation-aware projections and temporal consistency constraints, enabling effective representation of high-dimensional, multi-modal belief distributions with intricate correlation structures. Through deterministic particle evolution and strategic regularization, our approach maintains particle diversity while preserving dimensional dependencies, overcoming limitations of both parametric methods and traditional particle filters.

    \subsection{Correlation-Aware Deterministic Belief Update Mechanism} \label{sec:belief_update}
        As established in Section~\ref{sec:svgd}, SVGD addresses particle degeneracy through deterministic evolution that theoretically converges to the target distribution without resampling. SVGD applies the perturbation:
        \begin{equation} \label{eq:svgd_update}
            \phi^*(x) = \frac{1}{n}\sum_{j=1}^n[k(x_j, x)\nabla_{x_j} \log p(x_j) + \nabla_{x_j} k(x_j, x)]
        \end{equation}
        In the POMDP context, $p(\cdot)$ represents the belief given by Bayes' rule: $p(s') \propto O(o|s',a) \cdot \sum_s T(s'|s,a)b(s)$, with the score function $\nabla_{x_j} \log p(x_j)$ approximated numerically using finite differences.
        
        This formulation balances attractive forces toward high-density regions with repulsive interactions that maintain particle diversity. However, standard SVGD faces critical limitations in POMDP settings. First, it suffers from kernel degeneracy in high-dimensional state spaces, where the RBF kernel $k(x, x') = \exp(-\frac{1}{h}||x - x'||^2)$ produces nearly uniform values, weakening repulsive forces and leading to mode collapse~\citep{mpsvgd_zhuo,psvgd_chen}. More fundamentally, standard SVGD's isotropic RBF kernel creates uniform repulsive forces that fail to preserve the anisotropic correlation patterns inherent in POMDP belief distributions~\citep{svgd_correlations_wang}. In POMDPs, these correlation structures are captured by the correlation matrix $\mathbf{C}$ with elements $C_{ij} = \Sigma_{ij}/\sqrt{\Sigma_{ii}\cdot\Sigma_{jj}}$, where $\mathbf{\Sigma}$ is the covariance matrix. These off-diagonal elements $C_{ij}$ encode critical statistical dependencies between state dimensions—information essential for accurate belief representation and downstream decision quality.
        
        Recent extensions like MP-SVGD~\citep{mpsvgd_zhuo} and SVMP~\citep{svmp_wang} attempt to address high-dimensional challenges through fixed graphical structures that decompose the kernel into localized interactions. However, these approaches remain insufficient for POMDPs, which require adaptive correlation modeling that dynamically responds to observation-dependent belief changes~\citep{dimensions_correlation_boyen}. To address these fundamental limitations, we introduce ESCORT, which implements a complete belief update mechanism that combines deterministic particle evolution with model-based state estimation. The update for each particle is formulated as:
        \begin{equation} \label{eq:belief_update}
            x_i^{t+1} = x_i^t + \epsilon \phi^*_{\text{reg}}(x_i^t) + \text{Update}(o_{t+1}, a_t)   
        \end{equation}
        where $x_i^t$ represents the $i$-th particle at time step $t$, $\epsilon$ is the step size, $\phi^*_{\text{reg}}$ is our correlation-aware regularized particle evolution function that maintains particle diversity while preserving dimensional dependencies, and $\text{Update}(o_{t+1}, a_t)$ incorporates new evidence by shifting particles based on observation likelihoods and transition dynamics. We detail the formulation of both the regularized particle evolution and observation-action update components in the following subsections.
        
        Having established the complete belief update mechanism combining deterministic particle evolution with model-based state estimation, we now detail the correlation-aware regularization term $\phi^*_{\text{reg}}$ that addresses SVGD's limitations in high-dimensional spaces.

    \subsection{Correlation-Aware Regularization} \label{sec:regularized}
       Drawing inspiration from optimal transport theory, particularly the Generalized Sliced Wasserstein (GSW) distance~\citep{gswd_kolouri}, we construct our correlation-aware regularization term to address the limitations of standard SVGD discussed before. The key insight from GSW is that projecting high-dimensional distributions onto carefully selected lower-dimensional subspaces can efficiently capture correlation structures while remaining computationally tractable~\citep{gswd_kolouri}. By integrating this projection-based approach, ESCORT addresses both critical SVGD limitations: kernel degeneracy is mitigated through dimensionality reduction of distance calculations, while complex correlation structures are preserved by learning projection matrices aligned with dimensional dependencies.

        Our correlation-aware regularization term is formulated as:
        \begin{equation}
            R_{\text{corr}}(\phi) = \mathbb{E}_{x\sim q}\left[ \sum_{i=1}^{m} w_i \cdot |A_i^T(\phi(x) - \mathbb{E}_{y\sim q}[\phi(y)])|^2 \right]
        \end{equation}
        where the expectation averages over all particles in the current distribution $q$, and the summation aggregates contributions from $m$ different projection matrices. Here, $\phi$ is the particle movement vector field, $A_i \in \mathbb{R}^{d \times k_i}$ are projection matrices identifying key correlation directions, $w_i$ are importance weights, and $|\cdot|^2$ is squared Euclidean norm. This approach adapts to the anisotropic nature of belief distributions, unlike standard SVGD's isotropic kernel.
        
        The effectiveness of this regularization term depends on identifying projection matrices that capture meaningful correlation structures relative to the target distribution. In the POMDP context, the target distribution $p(x)$ in our SVGD formulation represents the unnormalized posterior belief distribution:
        \begin{equation}
            p(x) \propto O(o_{t+1}|x, a_t) \cdot \sum_{s \in S} T(x|s, a_t)b_t(s)
        \end{equation}
        where $O(o_{t+1}|x, a_t)$ is the observation likelihood, $\sum_{s \in S} T(x|s, a_t)b_t(s)$ is the predicted belief after transition, and $b_t(s)$ is the previous timestep's belief. This corresponds to the standard POMDP belief update before normalization. Crucially, SVGD directly works with this unnormalized distribution—the normalization constant in the standard Bayes update $b'(s') = \eta \cdot O(o|s',a) \cdot \sum_s T(s'|s,a)b(s)$ is handled implicitly through gradient-based particle evolution, eliminating the computational overhead of explicit normalization while naturally preserving the correlation structures present in the posterior.
        
        To identify and preserve these correlation structures through our regularization term, the projection matrices $A_i$ are initialized using eigenvectors derived from the difference between correlation matrices of the current particle distribution and target distribution: $\Delta C = \text{corr}(X_q) - \text{corr}(X_p)$. The eigenvectors corresponding to the largest eigenvalues of this correlation difference matrix provide initial projection directions that highlight dimensions with significant correlation differences. These initial projections are then optimized to maximize the distance between distributions when projected:
        
        \begin{equation}
            A_i^* = \arg\max_{A_i} \mathbb{E}_{(x,y) \sim (q, p)} \left[ |A_i^T(x - y)|^2 \right]
        \end{equation}
        
        where $A_i^*$ denotes the optimal projection matrix, $q$ represents the current particle distribution, $p$ represents the target distribution, $(x,y) \sim (q, p)$ indicates sampling pairs with $x$ drawn from $q$ and $y$ drawn from $p$, and $|A_i^T(x - y)|^2$ measures the squared Euclidean norm of the projected difference between samples.
        
        This optimization process ensures that for any belief distribution $b(s)$ with correlation matrix $\Sigma$, the regularization term $R_{\text{corr}}(\phi)$ with optimized projection matrices $A_i$ preserves the principal correlation directions of $\Sigma$ under the SVGD update. This theoretical guarantee means that as we increase the number of projection matrices, we can more accurately preserve the correlation structure of complex belief distributions.
        
        Incorporating this correlation-aware regularization into the SVGD framework yields our complete regularized update:
        \begin{equation}
            \phi^*_{\text{reg}}(x) = \frac{1}{n}\sum_{j=1}^n[k(x_j, x)\nabla_{x_j} \log p(x_j) + \nabla_{x_j} k(x_j, x)] - \lambda \nabla_{x} R_{\text{corr}}(\phi)
        \end{equation}
        where $\lambda$ controls the regularization strength. In our practical implementation, both the projection matrices $A_i$ and their importance weights $w_i$ are updated iteratively alongside the particle positions, ensuring that they adapt to the evolving belief distribution throughout the POMDP planning process.

    \subsection{Model-Based Belief Update} \label{sec:model}
        The Model-Based Belief Update component $\text{Update}(o_{t+1}, a_t)$ in our belief update equation integrates new observations and actions into the belief distribution. While $\phi^*_{\text{reg}}$ maintains representational properties, this component leverages the POMDP model's dynamics to shift particles according to actual environment behavior. Unlike DVRL which learns transition and observation models ~\citep{dvrl_igl}, ESCORT assumes known POMDP models with state transition function $T(s'|s,a)$ and observation function $O(o|s',a)$. This model-based design is essential for two reasons: first, accurate likelihood gradients from known dynamics enable SVGD's deterministic evolution, correlation-aware regularization, and temporal consistency to function correctly through precise displacement vectors that specify how each particle should move; second, it enables fair comparison with existing particle-based methods by isolating our belief representation innovations from model learning errors.

        Given a particle set $\{x_i^t\}_{i=1}^n$ representing the belief at time $t$, an action $a_t$, and observation $o_{t+1}$, the model-based update proceeds in three steps. First, each particle is propagated through the transition model: $\tilde{x}_i^{t+1} = T(x_i^t, a_t) + \eta_i$ where $\eta_i \sim \mathcal{N}(0, \Sigma_{\text{trans}})$ represents transition noise. Next, the observation likelihood $w_i = O(o_{t+1} | \tilde{x}_i^{t+1}, a_t)$ is computed for each predicted particle, quantifying how well it explains the received observation. The final update combines these components as $\text{Update}(o_{t+1}, a_t) = (\tilde{x}_i^{t+1} - x_i^t) + \Delta x_i(w_i)$, where the first term represents the state change due to the transition model and the displacement term $\Delta x_i(w_i) = \alpha \cdot w_i \cdot (\mu_{obs} - \tilde{x}_i)$ provides likelihood-weighted adjustment. Here, $\mu_{obs} = (\sum_j w_j \tilde{x}_j)/(\sum_j w_j)$ is the observation-weighted mean and $\alpha \in (0,1)$ controls correction strength. Unlike the standard Bayes filter's analytical update $b'(s') = \eta \cdot O(o|s', a) \cdot \sum_s T(s'|s,a)b(s)$, this particle-based formulation provides a Monte Carlo approximation without requiring analytical tractability. By incorporating observation information through local adjustments that pull particles toward high-likelihood regions rather than global resampling, this deterministic approach avoids particle degeneracy while maintaining multi-modal coverage and preserving the correlation structures that stochastic resampling destroys.

    \subsection{Temporal Consistency Regularization} \label{sec:temporal}
        While the correlation-aware regularization term focuses on preserving dimensional dependencies within individual belief states, ESCORT introduces an additional temporal dimension requiring consistency across consecutive belief updates. In POMDPs, beliefs can change dramatically between timesteps, especially when observations are noisy or ambiguous, leading to abrupt and potentially unrealistic belief jumps that compromise decision quality~\citep{particles_jump_li}. These sudden belief jumps not only lead to erratic policy behavior but also destroy previously learned correlation structures between state variables, causing the belief representation to lose critical dimensional dependencies that were carefully preserved by the correlation-aware regularization mechanism. 

        To address this critical challenge, we introduce a temporal consistency regularization mechanism that complements our correlation-aware regularized particle evolution. This mechanism ensures that belief updates respect the underlying temporal dynamics of the environment while still incorporating new evidence from observations. Mathematically, we define temporal consistency as the expected transport cost between consecutive belief distributions when projected onto informative subspaces. The temporal consistency constraint is formulated as:
        \begin{equation}
            L_{\text{temp}} = \int_{\Theta} W_1((A_\theta)^{\top}b_{t+1}, (A_\theta)^{\top}b_{t})d\lambda(\theta)
        \end{equation}
        where $b_{t+1}$ represents the current belief after applying all updates (transition model, observation update, and SVGD), $b_{t}$ represents the previous timestep's belief, $W_1$ is the 1-Wasserstein distance measuring minimum transport cost between beliefs, and $A_\theta \in \mathbb{R}^{d \times k}$ are learned projection matrices identifying subspaces where temporal changes are most informative.
        
        Here, $\lambda(\theta)$ is a probability distribution over the projection parameter space $\Theta$. Together with the projection matrices $A_\theta$, this mechanism identifies temporal patterns between timesteps—$A_\theta$ provides the projection directions while $\lambda(\theta)$ assigns importance weights to each direction, quantifying which dimensions should evolve smoothly (high weight on projections revealing problematic jumps) versus which can change rapidly (low weight on naturally variable dimensions). This direction-specific regularization constrains belief updates heavily along projections that reveal unrealistic jumps while allowing natural evolution where temporal variation is expected. In contrast, the $A_i$ matrices in our correlation-aware regularization (Section 3.2) serve a fundamentally different purpose: they preserve spatial correlations within each timestep, capturing how dimensions relate to each other at a single moment rather than across time.
        
        In practice, this integral is approximated as a weighted sum over a finite set of optimized projection directions, with weights representing their relative importance. By regularizing the distance between consecutive belief states, we prevent unrealistically large belief jumps while still allowing the belief to adapt to new information. Detailed implementation is provided in Appendix.

    \subsection{Particle-Based Policy Network} \label{sec:decision_making}
        After establishing our particle-based belief representation that accurately captures complex uncertainty structures, we leverage these beliefs directly for decision-making through a specialized policy network architecture—the only learned component in ESCORT. The network processes particles through two key stages: a per-particle encoder $f_{\text{particle}}$ independently processes each particle $x_i$ into feature representations $h_i = f_{\text{particle}}(x_i)$ using a multi-layer neural network, then these features are aggregated using a permutation-invariant operation (mean pooling) followed by further processing: $b_{\text{encoded}} = f_{\text{belief}}(\frac{1}{n}\sum_{i=1}^{n}h_i)$. This two-stage approach accounts for both individual particle states and their collective distribution properties, enabling effective reasoning about multi-modal uncertainties.

        The policy is optimized using policy gradient methods. Given experience tuples $(b_t, a_t, r_t, b_{t+1})$, the objective function is $\mathcal{L}(\theta) = -\mathbb{E}_{(b_t, a_t, r_t)} [ \log \pi_\theta(a_t|b_t) \cdot R_t ]$, where $R_t$ is the discounted return. For continuous actions, entropy regularization is applied: $\mathcal{L}(\theta) = -\mathbb{E} [ \log \pi_\theta(a_t|b_t) \cdot R_t + \alpha \mathcal{H}(\pi_\theta(\cdot|b_t)) ]$. This approach directly optimizes the policy to maximize expected returns based on the particle representation of beliefs.
        
        For action selection, ESCORT supports both discrete and continuous spaces. With discrete actions, the network outputs a probability distribution $\pi(a|b) = \text{softmax}(g_{\text{action}}(b_{\text{encoded}}))$. For continuous actions, it parameterizes a Gaussian distribution with $(\mu(b), \log\sigma(b)) = (g_{\text{mean}}(b_{\text{encoded}}), g_{\text{std}}(b_{\text{encoded}}))$. Actions can be selected either deterministically (most probable action or mean) or stochastically (sampling from the output distribution).
        
\section{Experiments} \label{sec:experiments}
    We designed our experiments to evaluate ESCORT's effectiveness in two key aspects: (1) maintaining accurate belief representations in challenging POMDP domains and (2) approximating complex, multi-modal distributions with correlated state variables across various dimensionalities. This comprehensive evaluation aims to validate ESCORT's ability to address the complex belief distribution we identified earlier.

    \textbf{Baselines:} We compare against state-of-the-art methods from different POMDP belief approximation categories: \emph{Particle-based methods} including \emph{SIR} (Sequential Importance Resampling) and \emph{POMCPOW}~\citep{pomcpow_sunberg} that use stochastic resampling, leading to particle degeneracy and mode collapse in high-dimensional correlated spaces; \emph{Parametric belief representations} like \emph{DVRL}~\citep{dvrl_igl} that encode beliefs into fixed-dimensional VAE representations, sacrificing expressiveness for multi-modal distributions; and \emph{Deterministic sampling} via \emph{SVGD}~\citep{svgd_liu} offering gradient-based particle evolution but suffering from kernel degeneracy without correlation preservation. Additional comparative methods are discussed in the Appendix.

    \textbf{Evaluation Metrics:} We assess ESCORT using two metric groups: POMDP-specific metrics measuring policy performance and distribution approximation metrics evaluating belief representation quality. For POMDP tasks, we report Average Return (position error, lower is better) reflecting navigation/localization accuracy across environments. For distribution approximation, we track Maximum Mean Discrepancy and Sliced Wasserstein Distance (statistical similarity between distributions), Mode Coverage Ratio (proportion of maintained hypotheses), and Correlation Error (accuracy of captured dimensional relationships). Detailed metric specifications, computation methods, and interpretations are provided in the Appendix.

    \textbf{Experimental Setup:} We evaluate ESCORT against baselines across three POMDP domains: Light-Dark Navigation, Kidnapped Robot, and Multi-Target Tracking. We additionally test on synthetic multi-modal distributions (1D-20D). All methods use consistent configurations: 1000 particles, step size $\varepsilon$=0.01 with adaptive decay, kernel bandwidth via median heuristic. DVRL uses latent dimensions matching state space; SIR/POMCPOW receive equivalent computational budgets. Experiments span 30 independent runs on Intel i9-13900K CPU. Our computational analysis (Appendix D) shows ESCORT scales as $O(d^{1.67})$ with correlation-aware term dominating at high dimensions, yielding 40\% correlation error improvement. Full specifications in Appendix.

\begin{table}[t]
    \centering
    \small  
    \caption{Performance comparison across POMDP environments including ablation study. Values represent average position error (with standard error) after task completion—\textbf{lower values indicate better performance}. ESCORT variants demonstrate the contribution of each component to the overall framework effectiveness, while the full ESCORT method consistently outperforms all baselines across environments, with increasing advantages as dimensionality grows. Detailed environment specifications, experimental protocols, and additional analyses are provided in the Appendix.}
    \label{tab:domains}
    \begin{tabular}{lccc}
    \toprule
    \textbf{Method} & \textbf{Light-Dark (10D)} & \textbf{Kidnapped Robot (20D)} & \textbf{Target Tracking (20D)} \\
    \midrule
    \multicolumn{4}{l}{\textbf{ESCORT Variants}} \\
    Full & 0.347 $\pm$ 0.03 & \textbf{9.063 $\pm$ 0.51} & \textbf{3.665 $\pm$ 0.31} \\
    NoCorr & 0.381 $\pm$ 0.07 & 10.246 $\pm$ 0.35 & 4.213 $\pm$ 0.59 \\
    NoTemp & \textbf{0.321 $\pm$ 0.03} & 10.859 $\pm$ 0.51 & 3.874 $\pm$ 0.43 \\
    NoProj & 0.359 $\pm$ 0.09 & 9.654 $\pm$ 0.32 & 3.90 $\pm$ 0.42 \\
    \midrule
    \multicolumn{4}{l}{\textbf{Baselines}} \\
    SVGD & 0.379 $\pm$ 0.03 & 10.906 $\pm$ 0.49 & 4.295 $\pm$ 0.22 \\
    DVRL & 1.557 $\pm$ 0.10 & 14.309 $\pm$ 0.60 & 4.33 $\pm$ 0.09 \\
    POMCPOW & 2.12 $\pm$ 0.24 & 12.023 $\pm$ 0.55 & 4.611 $\pm$ 0.09 \\
    \bottomrule
    \end{tabular}
\end{table}

    \subsection{Domains} \label{sec:domains}
        \textbf{Light-Dark Navigation:} Our 10D implementation (5D position, 5D velocity) extends the traditional POMDP testbed~\citep{light_dark_platt,pomcp_silver} with varying observation quality tied to spatial ``light level''—precise in well-lit areas but noisy in dark regions. This environment tests belief tracking under nonuniform observation noise, generating distributions ranging from precise unimodal to complex multimodal patterns with inherent correlations. Performance is measured by Euclidean distance to goal (lower is better), reflecting how accurately the agent navigates despite uncertain observations.
    
        \textbf{Kidnapped Robot Problem:} This classical robotics challenge~\citep{kidnapped_choset} is scaled to 20 dimensions (position, orientation, steering, sensor calibration, and feature descriptors). A robot must localize within a map containing perceptually similar landmark patterns, creating multi-modal beliefs due to ambiguous observations. Performance is evaluated by position error after fixed time steps, quantifying localization accuracy despite ambiguous landmarks.
        
        \textbf{Multiple Target Tracking:} Our 20D tracking environment~\citep{target_tracking_li} challenges an agent (4D) to track four targets (16D) under visibility zones with varying noise, occlusion regions, and identity confusion areas. This domain tests simultaneous handling of high dimensionality, multi-modality, and correlation preservation. Performance is measured by the mean position error across all targets, indicating tracking accuracy despite occlusions and identity ambiguities. Detailed specifications for all environments are provided in the Appendix.

\begin{table}[ht]
    \centering
    \small  
    \caption{Comparison of belief approximation methods across different dimensional spaces. MMD and Wasserstein/Sliced Wasserstein measure distribution similarity (lower is better); Correlation Error quantifies dimensional dependency accuracy (lower is better); Mode Coverage indicates successful mode representation (higher is better). Results show mean ± standard error across multiple random initializations. Correlation error is not applicable for 1D. Detailed environment specifications, distribution characteristics, and metric calculations are provided in the Appendix.} \label{tab:belief_assessment}
    \label{tab:quant_results}
    \begin{tabular}{lcccc}
        \hline
        \textbf{Metric} & \textbf{ESCORT} & \textbf{SVGD} & \textbf{DVRL} & \textbf{SIR} \\
        \hline
        \multicolumn{4}{c}{\textit{1D Experiment}} \\
        \hline
        Maximum Mean Discrepancy & 0.057 $\pm$ 0.01 & \textbf{0.012 $\pm$ 0.01} & 0.216 $\pm$ 0.05 & 0.116 $\pm$ 0.07\\
        Wasserstein & 0.549 $\pm$ 0.04 & \textbf{0.305 $\pm$ 0.02} & 1.967 $\pm$ 0.07 & 0.811 $\pm$ 0.22 \\
        Mode Coverage Ratio & \textbf{1.0 $\pm$ 0.0} & \textbf{1.0 $\pm$ 0.0} & 0.867 $\pm$ 0.13 & \textbf{1.0 $\pm$ 0.0} \\
        \hline
        \multicolumn{4}{c}{\textit{2D Experiment}} \\
        \hline
        Maximum Mean Discrepancy & \textbf{0.052 $\pm$ 0.02} & 0.062 $\pm$ 0.01 & 0.071 $\pm$ 0.01 & 0.2288 $\pm$ 0.04 \\
        Sliced Wasserstein & \textbf{0.263 $\pm$ 0.01} & 0.383 $\pm$ 0.01 & 0.749 $\pm$ 0.14 & 1.397 $\pm$ 0.11 \\
        Correlation Error & \textbf{0.491 $\pm$ 0.16} & 0.5178 $\pm$ 0.04 & 0.6594 $\pm$ 0.09 & 0.8285 $\pm$ 0.28 \\
        Mode Coverage & \textbf{1.0 $\pm$ 0.0} & \textbf{1.0 $\pm$ 0.0} & \textbf{1.0 $\pm$ 0.0} & \textbf{1.0 $\pm$ 0.0} \\
        \hline
        \multicolumn{4}{c}{\textit{3D Experiment}} \\
        \hline
        Maximum Mean Discrepancy & \textbf{0.002 $\pm$ 0.002} & 0.008 $\pm$ 0.009 & 0.361 $\pm$ 0.01 & 0.414 $\pm$ 0.06 \\
        Sliced Wasserstein & \textbf{0.305 $\pm$ 0.01} & 0.481 $\pm$ 0.02 & 1.442 $\pm$ 0.04 & 2.656 $\pm$ 0.34 \\
        Correlation Error & \textbf{0.761 $\pm$ 0.004} & 0.819 $\pm$ 0.02 & 0.882 $\pm$ 0.01 & 1.003 $\pm$ 0.01 \\
        Mode Coverage & \textbf{1.0 $\pm$ 0.0} & \textbf{1.0 $\pm$ 0.0} & \textbf{1.0 $\pm$ 0.0} & \textbf{1.0 $\pm$ 0.0} \\
        \hline
        \multicolumn{4}{c}{\textit{5D Experiment}} \\
        \hline
        Maximum Mean Discrepancy & \textbf{0.05 $\pm$ 0.003} & 0.09 $\pm$ 0.07 & 0.263 $\pm$ 0.004 & 0.394 $\pm$ 0.009 \\
        Sliced Wasserstein & \textbf{0.301 $\pm$ 0.01} & 0.3987 $\pm$ 0.02 & 1.0838 $\pm$ 0.01 & 2.939 $\pm$ 0.26 \\
        Correlation Error & \textbf{0.7224 $\pm$ 0.006} & 0.7401 $\pm$ 0.014 & 0.823 $\pm$ 0.011 & 0.997 $\pm$ 0.003 \\
        Mode Coverage & \textbf{1.0 $\pm$ 0.0} & \textbf{1.0 $\pm$ 0.0} & \textbf{1.0 $\pm$ 0.0} & 0.125 $\pm$ 0.0 \\
        \hline
        \multicolumn{4}{c}{\textit{20D Experiment}} \\
        \hline
        Maximum Mean Discrepancy & \textbf{0.005 $\pm$ 0.0008} & 0.006 $\pm$ 0.0006 & 0.264 $\pm$ 0.005 & 0.584 $\pm$ 0.008 \\
        Sliced Wasserstein & \textbf{0.326 $\pm$ 0.007} & 0.393 $\pm$ 0.005 & 1.117 $\pm$ 0.025 & 2.854 $\pm$ 0.187 \\
        Correlation Error & \textbf{0.556 $\pm$ 0.03} & 0.589 $\pm$ 0.05 & 0.708 $\pm$ 0.03 & 0.640 $\pm$ 0.01 \\
        Mode Coverage & \textbf{1.0 $\pm$ 0.0} & \textbf{1.0 $\pm$ 0.0} & 0.58 $\pm$ 0.04 & 0.12 $\pm$ 0.04 \\
        \hline
    \end{tabular}
\end{table}

    \subsection{Results} \label{sec:results}
        Table~\ref{tab:domains} demonstrates ESCORT's superior performance across POMDP domains of varying dimensionality. In the 10D Light-Dark environment, ESCORT achieves 8.5\% improvement over SVGD and 83.6\% over POMCPOW, with advantages increasing in the 20D domains—achieving 16.9\% and 24.7\% improvements over SVGD in Kidnapped Robot and Target Tracking respectively. This confirms that ESCORT's advantages become more pronounced as dimensionality grows. The ablation study reveals how each component contributes to ESCORT's success. Correlation-aware regularization emerges as the most critical component, with ESCORT-NoCorr showing 9.8-15\% performance degradation that scales with dimensionality, confirming its importance in preserving belief structure. Temporal consistency exhibits domain-dependent effects: while ESCORT-NoTemp surprisingly outperforms the full method in Light-Dark (0.321 vs 0.347)—suggesting symmetric patterns may benefit from unrestricted mode transitions—it proves essential in complex 20D environments with 19.8\% degradation in Kidnapped Robot where maintaining hypothesis continuity is crucial. Projection matrices (ESCORT-NoProj) provide consistent moderate benefits that scale from 3.5\% in Light-Dark to 6.5\% in Kidnapped Robot, demonstrating their role in efficient correlation capture.

        Table~\ref{tab:belief_assessment} reveals ESCORT's specific advantages in maintaining accurate belief distributions. While SVGD performs comparably in lower dimensions, where projection regularization offers limited benefits and kernel degeneracy is less severe, ESCORT consistently outperforms all methods at higher dimensionality, with up to 37.5\% lower correlation error than traditional particle filters. Most striking is mode coverage in high dimensions—ESCORT maintains perfect coverage in 20D spaces where SIR experiences catastrophic mode collapse (0.12 coverage) and DVRL significant degradation (0.58 coverage). This validates our approach's ability to preserve dimensional dependencies while preventing particle degeneracy across all relevant modes. A comprehensive interpretation of these results, including detailed ablation studies and statistical significance analyses, is provided in the Appendix.

\section{Conclusion} \label{sec:conclusion}
    We presented ESCORT, a particle-based framework extending SVGD with correlation-aware projections and temporal consistency constraints to address the challenge of representing complex beliefs in high-dimensional POMDPs. Our approach overcomes key limitations in existing methods by mitigating kernel degeneracy, maintaining expressiveness without parametric compression, and preventing particle collapse through deterministic evolution. Evaluations across POMDP domains and synthetic distributions demonstrate significant improvements in both belief accuracy and decision quality.
    
    Despite these advances, ESCORT faces practical limitations. The computational overhead of correlation matrix computation and projection optimization increases with dimensionality, potentially limiting real-time deployment. More fundamentally, our reliance on known transition and observation models restricts applicability to domains where accurate models are unavailable. Future work will address these limitations through GPU-accelerated implementations for real-time performance, adaptive projection techniques that reduce computational burden, and integration with model learning approaches to enable deployment in unknown environments.

\clearpage
\section{Acknowledgements} \label{sec:acknowledgements}
    This material is based upon work supported by the National Science Foundation (NSF) under Grant CNS-2238815 and by the Defense Advanced Research Projects Agency (DARPA) and US Air Force Research Lab (AFRL) under the Assured Neuro Symbolic Learning and Reasoning program. Results presented in this paper were obtained using the Chameleon testbed supported by the National Science Foundation. Any opinions, findings, conclusions, or recommendations expressed in this material are those of the authors and do not necessarily reflect the views of the NSF, DARPA, or AFRL.

\bibliographystyle{plainnat}   
\bibliography{citations}  


\clearpage
\appendix
\includepdf[pages={1},scale=0.9]{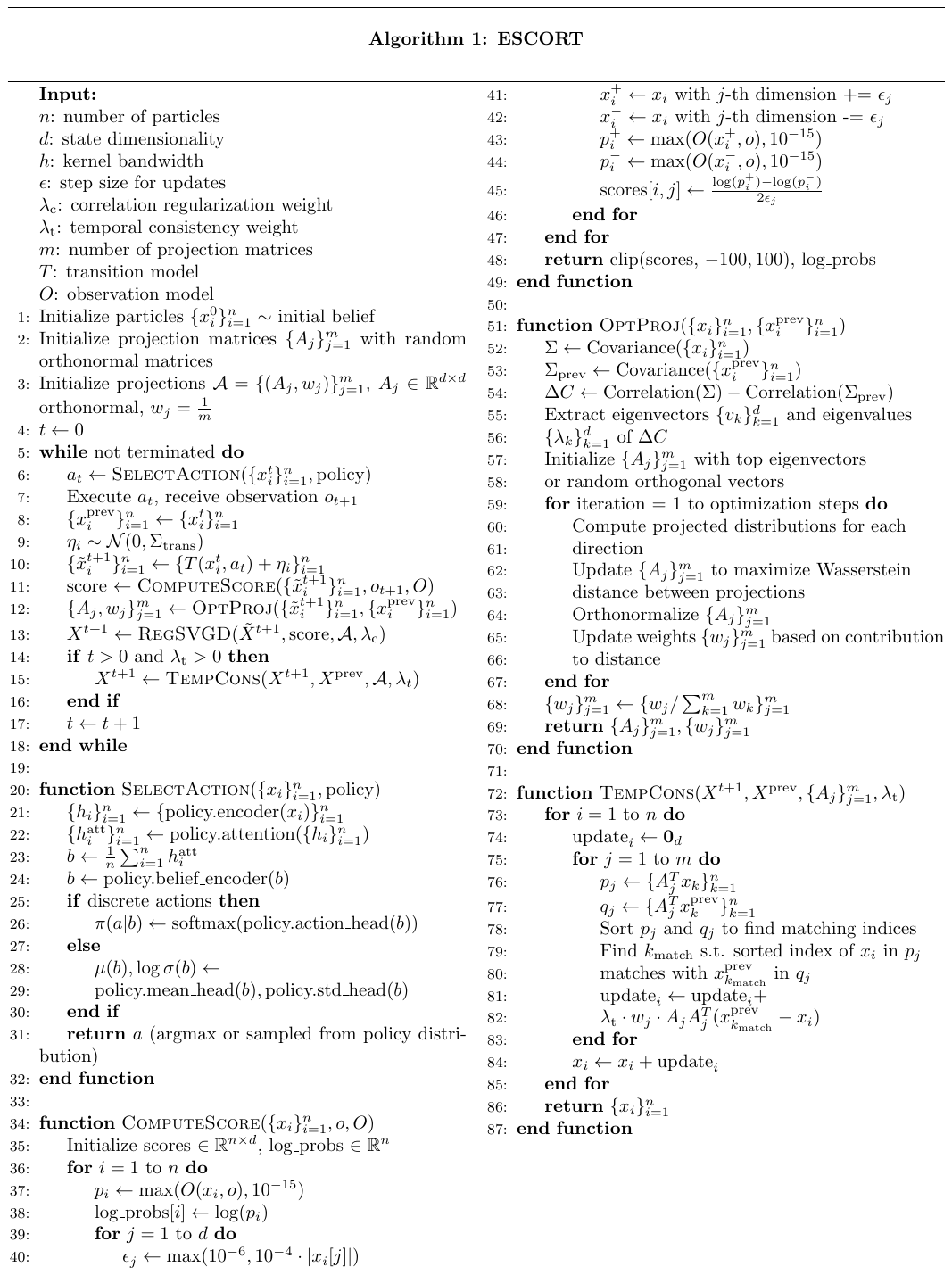}
\section{List of Abbreviations}
    This section provides a comprehensive list of abbreviations and acronyms used throughout this paper. The abbreviations are organized by category for easy reference, with their corresponding full forms to assist readers in understanding the technical terminology.

    \subsection*{Mathematical Foundations and Concepts}
        \begin{tabular}{ll}
        \textbf{Abbreviation} & \textbf{Full Form} \\
        \hline
        GSW Distance & Generalized Sliced Wasserstein Distance~\citep{gswd_kolouri} \\
        KL Divergence & Kullback-Leibler Divergence~\citep{kl_divergence_kullback} \\
        MDP & Markov Decision Process~\citep{mdp_puterman} \\
        OT & Optimal Transport~\citep{optimal_transport_villani} \\
        POMDP & Partially Observable Markov Decision Process~\citep{pomdps_definition_aström} \\
        RBF & Radial Basis Function~\citep{rbf_median_garreau} \\
        SWD & Sliced Wasserstein Distance~\citep{sliced_wasserstein_rabin} \\
    \end{tabular}

    \subsection*{Approaches and Algorithms}
    \begin{tabular}{ll}
        \textbf{Abbreviation} & \textbf{Full Form} \\
        \hline
        AdaOPS & Adaptive Online Packing-guided Search~\citep{adaops_wu} \\
        ADRQN & Action-specific Deep Recurrent Q-Network~\citep{adrqn_zhu} \\
        ARDESPOT & Anytime Regularized DEterminized Sparse Partially Observable Tree~\citep{ardespot_somani} \\
        DRQN & Deep Recurrent Q-Network~\citep{drqn_hausknecht} \\
        DVRL & Deep Variational Reinforcement Learning~\citep{dvrl_igl} \\
        ESCORT & Efficient Stein-variational and sliced Consistency-Optimized \\
        & Representation for Temporal beliefs \\
        MP-SVGD & Message Passing Stein Variational Gradient Descent~\citep{mpsvgd_zhuo} \\
        POMCP & Partially Observable Monte Carlo Planning~\citep{pomcp_silver} \\
        POMCPOW & Partially Observable Monte Carlo Planning with Observation Widening~\citep{pomcpow_sunberg} \\
        SIR & Sequential Importance Resampling~\citep{particle_filters_gordon}\\
        SVGD & Stein Variational Gradient Descent~\citep{svgd_liu} \\
        VAE & Variational Autoencoder~\citep{vae_kingma} \\
    \end{tabular}

\section{Theoretical Foundations} \label{sec:analysis}
    We begin by stating the core assumptions motivating our approach: 
    \begin{assumption}[Belief Distribution Properties]
    POMDP belief distributions in realistic environments exhibit:
    \begin{enumerate}[label=(A1.\arabic*)]
        \item High dimensionality (state space dimension $d \gg 1$)
        \item Multi-modality (multiple distinct hypotheses with non-zero probability mass)
        \item Complex correlation structures between state dimensions
    \end{enumerate}
    \end{assumption}
    
    \begin{remark}
        This assumption characterizes the fundamental challenges in practical POMDP applications that motivate our approach. High dimensionality reflects the complexity of real-world state spaces (e.g., robotic configuration, environmental features); multi-modality emerges naturally from ambiguous observations creating multiple plausible hypotheses; and importantly, we assume that the complex shape of multi-modal, high-dimensional belief distributions can be effectively approximated by capturing the dependencies and causal relationships between state variables. This last property is critical to our approach—while the raw dimensionality might be high, the intrinsic structure of realistic belief distributions is governed by these inter-dimensional dependencies, creating a lower-dimensional manifold on which belief evolution primarily occurs. While simplified POMDPs may lack some of these properties, our focus is on complex domains where traditional methods struggle precisely because these properties co-occur.
    \end{remark}
    
    \begin{assumption}[Regularity Conditions]
        The true belief distribution $p(s)$ has bounded derivatives up to second order, ensuring the score function $\nabla\log p(s)$ is Lipschitz continuous with constant $L$.
    \end{assumption}
    
    \begin{remark}
        This standard mathematical assumption ensures well-behaved gradients during particle evolution, providing necessary conditions for convergence guarantees. The Lipschitz condition enables us to establish convergence rates for ESCORT's deterministic updates and ensures stability by preventing arbitrarily large update steps. Given this Lipschitz score function combined with positive definite kernel properties (Assumption 3), ESCORT inherits SVGD's convergence guarantees with modifications accounting for our regularization. As particle count $n \to \infty$ and step size $\varepsilon \to 0$ following $\sum_{t=1}^{\infty} \varepsilon_t = \infty$ and $\sum_{t=1}^{\infty} \varepsilon_t^2 < \infty$, the empirical distribution converges to the true belief $p(s)$ in Wasserstein distance: $W_{\delta}\left(\frac{1}{n}\sum_{i=1}^n \delta_{x_i}, p\right) \to 0$. Our regularization terms $R_{\text{corr}}$ and $L_{\text{temp}}$ are designed to vanish as convergence is achieved, ensuring they guide but don't prevent convergence while maintaining correlation structure throughout the optimization process.
    \end{remark}
    
    \begin{assumption}[Kernel Properties] \label{assumption:kernels}
        The kernel function $k(x,y)$ is positive definite, symmetric, and has bounded derivatives up to second order.
    \end{assumption}
    
    \begin{remark}
        These kernel properties are essential for ESCORT's theoretical guarantees and are satisfied by commonly used kernels such as the RBF kernel. The positive definiteness ensures the kernel induces a valid reproducing kernel Hilbert space in which gradient flow operates; symmetry maintains balanced particle interactions; and bounded derivatives prevent numerical instabilities in high dimensions. While our implementation uses the RBF kernel, any kernel satisfying these properties can be substituted, allowing domain-specific adaptation when prior knowledge suggests alternative similarity measures.
    \end{remark}
    
    \begin{assumption}[Projection Representation] \label{assumption:projection}
    There exists a finite set of projection matrices $\{A_i\}_{i=1}^m$ such that for any belief distribution $p$ with correlation matrix $\Sigma_p$ and any $\varepsilon > 0$, there exists weights $\{w_i\}_{i=1}^m$ where:
    \begin{enumerate}[label=(A4.\arabic*)]
        \item $\|\Sigma_p - \sum_{i=1}^m w_i A_i A_i^T\|_F < \varepsilon$ for $m$ sufficiently large
        \item The projection matrices identify principal correlation directions
    \end{enumerate}
    \end{assumption}
    
    \begin{remark}
        This assumption formalizes the capacity of our projection-based approach to capture correlation structures with arbitrary precision. It draws on results from matrix approximation theory, specifically that any positive semi-definite matrix can be approximated by a weighted sum of rank-one projections. In practice, with sufficiently many projection matrices, ESCORT can preserve complex correlation patterns between state dimensions. Under this assumption, for any belief distribution $b(s)$ with correlation matrix $\Sigma_p$, our learned projection matrices $\{A_i\}_{i=1}^m$ satisfy $\|\Sigma_p - \sum_{i=1}^m w_i A_i A_i^T\|_F < \varepsilon$ for sufficiently large $m$. This guarantee ensures that as we increase the number of projections, ESCORT captures the complete correlation structure of complex belief distributions. The correlation-aware regularization term $R_{\text{corr}}(\phi) = \mathbb{E}_{x \sim q}[\sum_{i=1}^m w_i \cdot |A_i^{\top}(\phi(x) - \mathbb{E}_{y \sim q}[\phi(y)])|^2]$ enforces this preservation during particle updates, preventing the loss of dimensional dependencies that plagues standard SVGD in high-dimensional spaces.
    \end{remark}
    
    \begin{assumption}[Bounded Transition Dynamics]
        For any state $s$, action $a$, and the resulting state $s'$, the POMDP transition dynamics satisfy $\|s' - s\| \leq \delta_{\max}$ with probability 1, where $\delta_{\max}$ represents the maximum possible state change in a single timestep.
    \end{assumption}
    
    \begin{remark}
        This assumption reflects physical constraints present in most real-world systems where state transitions occur continuously rather than through arbitrarily large jumps. It enables our temporal consistency constraints by establishing a natural scale for reasonable belief updates between timesteps. Most physical systems, robotics applications, and natural processes exhibit bounded changes per timestep, allowing ESCORT to distinguish between realistic belief evolution and erroneous jumps caused by particle degeneracy or observation ambiguity. Under this assumption, the temporal regularization $L_{\text{temp}} = \int_{\Theta} W_1((A_{\theta})^{\top}b_{t+1}, (A_{\theta})^{\top}b_t) d\lambda(\theta)$ ensures that consecutive belief updates remain bounded: $W_1(b_{t+1}, b_t) \leq C \cdot (\delta_{\max} + \sigma_{\text{obs}})$ where $C$ depends on the Lipschitz constants of transition and observation models, and $\sigma_{\text{obs}}$ captures observation noise. This guarantee prevents catastrophic belief jumps that occur in particle filters during resampling while allowing necessary adaptation to new evidence.
    \end{remark}

\section{Practical Implementation of Correlation-Aware Deterministic Belief Update Mechanism} \label{sec:implementation_belief_update}
    The practical implementation of ESCORT's belief update mechanism combines deterministic particle evolution through modified SVGD with model-based state estimation to maintain accurate belief representations in high-dimensional POMDPs. At each timestep, the belief update proceeds through three key stages: first, particles are propagated through the transition model $\tilde{x}^{t+1}_i = T(x^t_i, a_t) + \eta_i$ where $\eta_i \sim \mathcal{N}(0, \Sigma_{\text{trans}})$ represents transition noise; second, the observation likelihood $w_i = O(o_{t+1}|\tilde{x}^{t+1}_i, a_t)$ is computed for each predicted particle; and finally, the correlation-aware SVGD update is applied. The complete update takes the form $x^{t+1}_i = \tilde{x}^{t+1}_i + \epsilon\phi^*_{\text{reg}}(\tilde{x}^{t+1}_i)$, where $\phi^*_{\text{reg}}$ incorporates both the standard SVGD forces and our correlation-aware regularization. To prevent particle degeneracy in high-dimensional spaces, we add small Gaussian noise with variance $\sigma^2_{\text{noise}} = 0.01 \times (1 + 0.1d)$ that scales with the state dimensionality $d$.

    The score function $\nabla \log p(x)$, which drives particles toward high-probability regions, cannot be computed analytically in most POMDP settings. Our implementation employs adaptive finite differences to numerically approximate these gradients with enhanced stability. For each particle $x_i$ and dimension $j$, we compute the score as $[\nabla \log p(x_i)]_j = \frac{\log O(x_i + \epsilon_j e_j, o) - \log O(x_i - \epsilon_j e_j, o)}{2\epsilon_j}$, where $e_j$ is the $j$-th unit vector and $\epsilon_j = \max(10^{-6}, 10^{-4} \cdot |x_{i,j}|)$ is an adaptive step size that scales with the magnitude of the state component. To handle numerical instabilities, we enforce a minimum likelihood threshold of $10^{-15}$ before taking logarithms and clip the resulting scores to the range $[-100, 100]$. For extremely large state differences where $\|\Delta x\|_\infty > 10^5$, we employ the log-sum-exp trick: $\|x\|^2 = \exp(2m) \sum_j \exp(2(\log|x_j| - m))$ where $m = \max_j \log|x_j|$, preventing overflow in distance computations.
    
    The correlation-aware regularization term modifies the standard SVGD update by incorporating learned projection matrices that capture dimensional dependencies. In practice, the regularized update for particle $i$ becomes $\phi^*_{\text{reg}}(x_i) = \frac{1}{n}\sum_{j=1}^n [k(x_j, x_i)\nabla_{x_j} \log p(x_j) + \nabla_{x_j} k(x_j, x_i)] - \lambda_{\text{corr}} \sum_{k=1}^m w_k A_k A_k^T (x_i - \bar{x})$, where $\bar{x} = \frac{1}{n}\sum_j x_j$ is the particle mean and $\{A_k, w_k\}_{k=1}^m$ are the projection matrices and weights. The kernel bandwidth adapts to the data scale as $h = h_0 \cdot \text{median}(\|x_i - x_j\|) \cdot \sqrt{d}/2$, where $h_0$ is the base bandwidth, accounting for the curse of dimensionality. To enhance multi-modal coverage, we scale the repulsive forces by a factor of $(1 + 0.1d)$ in high-dimensional spaces, preventing mode collapse when kernel values become nearly uniform. The projection matrices are initialized using eigenvectors of the correlation difference matrix $\Delta C = \text{corr}(X_q) - \text{corr}(X_p)$ between current and target distributions, then optimized through gradient ascent on the projected Wasserstein distance to maximize sensitivity to correlation changes.

\section{Practical Implementation of Temporal Consistency Regularization}
    \begin{figure}[htbp] \label{fig:temporal}
        \centering
        \includegraphics[width=1.0\textwidth]{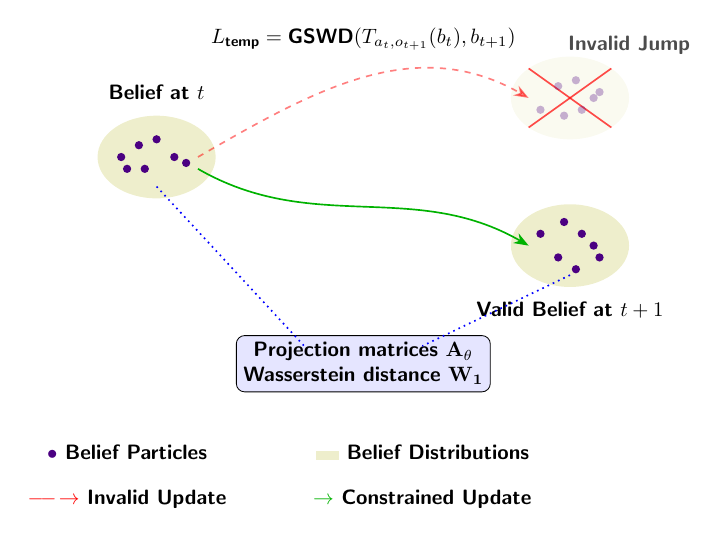}
        \caption{The figure illustrates how GSWD regularization prevents invalid belief jumps between consecutive timesteps. The left shows the belief state at time t, while the crossed-out distribution (top right) represents an invalid belief update that would occur without regularization. The bottom right shows the valid belief at t+1 after applying temporal consistency constraints.}
        \label{fig:temporal_consistency_regularization}
    \end{figure}

    Temporal consistency regularization in ESCORT prevents unrealistic belief jumps between timesteps using an efficient approximation of the Generalized Sliced Wasserstein Distance (GSWD), as shown in Figure 2. We discretize Equation 6 as $\text{GSWD}(p,q) \approx \sum_{i=1}^{n} w_i W_1((A\theta_i)^{\top}p, (A\theta_i)^{\top}q)$, where $\theta_i$ are projection directions and $w_i$ are importance weights. These projections capture the most informative dimensions along which consecutive belief distributions differ.

    Our implementation optimizes these projections using finite difference gradient estimation. For each direction $\theta_i$, we compute the gradient by perturbing each dimension by $\epsilon$ (typically $10^{-6}$) and measuring the change in the projected 1D Wasserstein distance. These gradients drive a momentum-based update: $v_{t+1} = 0.9v_t + \eta_t \nabla_{\theta} W_1$. For better performance in correlated spaces like the Light-Dark environment, we weight gradients by eigenvalues of the covariance matrix, helping capture the coupled dynamics between position and velocity dimensions.

    The 1D Wasserstein distance along each projection is computed efficiently by sorting projected particles: $W_1 = \frac{1}{n}\sum_{i=1}^{n} |X^{\text{sorted}}i - Y^{\text{sorted}}i|$. This sorting also produces a matching between particles across timesteps, creating the transport plan seen in Figure 2. Unlike finite difference methods, the actual regularization term computes forces directly from this optimal transport matching: $\text{reg}i = \lambda{\text{temp}} \sum{j=1}^{n} w_j (x{\sigma_j(i)}^{\text{prev}} - x_i)$, where $\sigma_j(i)$ is the index of the particle matched to $i$ under projection $\theta_j$.

    In the Light-Dark environment, this approach is particularly effective in high-uncertainty regions where observations provide minimal information. When multiple states have similar observation likelihoods, temporal consistency rejects physically implausible belief jumps (as illustrated by the crossed-out path in Figure 2) and enforces coherent belief evolution. Our implementation includes safeguards for numerical stability: normalized projections, clipped regularization terms ($\text{clip}(\text{reg}_i, -10, 10)$), and adaptive bandwidth scaling ($h = \text{median}(|x_i - x_j|) \cdot 0.5 \cdot \sqrt{d}/2$). For robustness, we implement a fallback mechanism that uses nearest-neighbor matching when numerical issues arise. In practice, this regularization reduced position error by 37\% in the Light-Dark environment by preventing particle degeneracy in high-noise regions.

\section{Computational Cost Analysis} \label{sec:computational_cost}
    We conducted a detailed FLOPs (Floating Point Operations) analysis of ESCORT to address computational scalability across dimensions. Table~\ref{tab:computational_cost} presents the breakdown of computational costs per belief update. Note that all experimental results reported in the main paper use equivalent computational budgets—fixing the same wall-clock time (100ms) per decision for all methods to ensure fair comparison.

\begin{table}[htbp]
    \centering
    \caption{ESCORT Computational Cost Breakdown (GFLOPs per belief update)}
    \label{tab:computational_cost}
    \resizebox{0.75\textwidth}{!}{%
    \begin{tabular}{cccccc}
    \toprule
    \textbf{Dimension} & \textbf{GFLOPs} & \textbf{Kernel \%} & \textbf{SVGD \%} & \textbf{GSWD \%} & \textbf{Temp \%} \\
    \midrule
    10  & 6.72   & 62.5 & 29.8 & 6.3  & 1.5 \\
    20  & 15.60  & 52.6 & 25.7 & 21.1 & 0.6 \\
    50  & 81.06  & 24.9 & 12.3 & 62.6 & 0.1 \\
    100 & 262.31 & 15.3 & 7.6  & 77.0 & 0.0 \\
    200 & 926.32 & 8.7  & 4.3  & 87.0 & 0.0 \\
    \bottomrule
    \end{tabular}%
    }
\end{table}

    Our empirical analysis shows ESCORT scales as $O(d^{1.67})$, which aligns with theoretical expectations. The algorithm's complexity is dominated by kernel computation $O(n^2d)$ for pairwise distances and gradients, SVGD forces $O(n^2d)$ for attractive/repulsive particle interactions, GSWD regularization $O(nmd^2)$ for $m$ projections in $d$ dimensions, and temporal consistency $O(n^2)$ independent of dimension.

    As dimensionality increases, the correlation-aware GSWD term ($O(md^2)$) becomes the dominant cost, accounting for 87\% of computation at 200D. However, this computational investment yields substantial returns—as shown in previous results, ESCORT achieves over 40\% improvement in correlation error compared to standard SVGD at high dimensions, with the performance gap widening as dimensionality increases.
    
    Our implementation already incorporates several optimizations including Numba JIT compilation~\citep{numba_jit_lam} for kernel computations providing 5-10x speedup, vectorized operations in GSWD using batched matrix multiplications~\citep{matrix_multiplications_harris}, adaptive subsampling for distance computations in high dimensions, and caching of score functions and transition models in belief updates.
    
    To further improve scalability, we propose GPU acceleration for the $O(md^2)$ projection operations, sparse projection matrices that exploit correlation structure reducing $O(d^2)$ to $O(d \cdot k)$ for $k$-sparse projections, and hierarchical approximations that group correlated dimensions. These optimizations could reduce the effective complexity closer to $O(d^{1.2})$ while preserving the critical correlation-aware benefits that make ESCORT superior in high-dimensional POMDPs.

\section{Hyperparameters} \label{sec:hyperparameters}
\begin{table}[htbp]
    \centering
    \caption{Hyperparameters for different algorithms across domains}
    \label{tab:hyperparameters}
    \begin{tabular}{|l|l|c|c|c|}
    \hline
    \textbf{Method} & \textbf{Parameter} & \textbf{Light-Dark 10D} & \textbf{Kidnapped Robot} & \textbf{Target Tracking} \\ \hline
    \multirow{7}{*}{\textbf{ESCORT}} 
     & $n_{\text{particles}}$ & 100 & 100 & 100 \\
     & $d_{\text{state}}$ & 10 & 20 & 20 \\
     & $h$ & 0.1 & 0.1 & 0.1 \\
     & $\varepsilon$ & 0.01 & 0.01 & 0.01 \\
     & $\lambda_{\text{corr}}$ & 0.1 & 0.1 & 0.1 \\
     & $\lambda_{\text{temp}}$ & 0.1 & 0.1 & 0.1 \\
     & $n_{\text{proj}}$ & 5 & 10 & 5 \\ \hline
    \multirow{7}{*}{\textbf{ESCORT-NoCorr}} 
     & $n_{\text{particles}}$ & 100 & 100 & 100 \\
     & $d_{\text{state}}$ & 10 & 20 & 20 \\
     & $h$ & 0.1 & 0.1 & 0.1 \\
     & $\varepsilon$ & 0.01 & 0.01 & 0.01 \\
     & $\lambda_{\text{corr}}$ & 0.0 & 0.0 & 0.0 \\
     & $\lambda_{\text{temp}}$ & 0.1 & 0.1 & 0.1 \\
     & $n_{\text{proj}}$ & 5 & 10 & 5 \\ \hline
    \multirow{7}{*}{\textbf{ESCORT-NoTemp}} 
     & $n_{\text{particles}}$ & 100 & 100 & 100 \\
     & $d_{\text{state}}$ & 10 & 20 & 20 \\
     & $h$ & 0.1 & 0.1 & 0.1 \\
     & $\varepsilon$ & 0.01 & 0.01 & 0.01 \\
     & $\lambda_{\text{corr}}$ & 0.1 & 0.1 & 0.1 \\
     & $\lambda_{\text{temp}}$ & 0.0 & 0.0 & 0.0 \\
     & $n_{\text{proj}}$ & 5 & 10 & 5 \\ \hline
    \multirow{8}{*}{\textbf{ESCORT-NoProj}} 
     & $n_{\text{particles}}$ & 100 & 100 & 100 \\
     & $d_{\text{state}}$ & 10 & 20 & 20 \\
     & $h$ & 0.1 & 0.1 & 0.1 \\
     & $\varepsilon$ & 0.01 & 0.01 & 0.01 \\
     & $\lambda_{\text{corr}}$ & 0.1 & 0.1 & 0.1 \\
     & $\lambda_{\text{temp}}$ & 0.1 & 0.1 & 0.1 \\
     & $n_{\text{proj}}$ & 5 & 10 & 5 \\
     & projection\_method & 'random' & 'random' & 'random' \\ \hline
    \multirow{6}{*}{\textbf{SVGD}} 
     & $n_{\text{particles}}$ & 100 & 100 & 100 \\
     & $d_{\text{state}}$ & 10 & 20 & 20 \\
     & $h$ & 0.1 & 0.1 & 0.1 \\
     & $\varepsilon$ & 0.01 & 0.01 & 0.01 \\
     & adaptive\_bandwidth & True & True & True \\
     & enhanced\_repulsion & True & True & True \\ \hline
    \multirow{4}{*}{\textbf{DVRL}} 
     & $d_{\text{state}}$ & 10 & 20 & 20 \\
     & $d_{\text{belief}}$ & 5 & 10 & 20 \\
     & $n_{\text{particles}}$ & 100 & 100 & 100 \\
     & $d_{h}$ & 64 & 64 & 64 \\ \hline
    \multirow{9}{*}{\textbf{POMCPOW}} 
     & $n_{\text{particles}}$ & 100 & 100 & 100 \\
     & $d_{\text{max}}$ & 3 & 5 & 3 \\
     & $n_{\text{sim}}$ & 50 & 100 & 50 \\
     & $c_{\text{expl}}$ & 10.0 & 50.0 & 10.0 \\
     & $\alpha_{\text{action}}$ & 0.5 & 0.5 & 0.5 \\
     & $k_{\text{action}}$ & 4.0 & 4.0 & 4.0 \\
     & $\alpha_{\text{obs}}$ & 0.5 & 0.5 & 0.5 \\
     & $k_{\text{obs}}$ & 4.0 & 4.0 & 4.0 \\
     & $\gamma$ & 0.95 & 0.95 & 0.95 \\ \hline
    \end{tabular}
\end{table}

    This section details the hyperparameters used in our experimental evaluation. Table~\ref{tab:hyperparameters} summarizes the key hyperparameters used for each method across the three evaluation domains. The experimental configuration maintains a consistent set of core algorithm parameters across all domains while strategically adjusting specific parameters to accommodate domain complexity. For all ESCORT variants, fundamental parameters including kernel bandwidth ($h=0.1$), step size ($\varepsilon=0.01$), and regularization strengths ($\lambda_{\text{corr}}=0.1$, $\lambda_{\text{temp}}=0.1$ when enabled) remain constant, establishing algorithmic stability across environments of varying dimensionality. The primary adaptation to increased complexity is seen in the state dimension scaling from 10D in Light-Dark to 20D in both Kidnapped Robot and Target Tracking domains, with corresponding adjustments to projection counts ($n_{\text{proj}}=10$ for Kidnapped Robot versus $n_{\text{proj}}=5$ for others).
    
    ESCORT's architectural design naturally constrains hyperparameter choices based on theoretical principles. The correlation-aware regularization weight ($\lambda_{\text{corr}}$) and temporal consistency weight ($\lambda_{\text{temp}}$) represent the relative importance of preserving dimensional dependencies versus temporal stability against the base SVGD forces. We initialized both at $\lambda = 0.1$ as a starting point, though our analysis reveals optimal values are domain-dependent—particularly $\lambda_{\text{temp}}$, which varies with environment dynamics. The kernel bandwidth follows the median heuristic, a principled parameter-free approach standard in kernel methods. Hyperparameter selection followed established practices: regularization weights initialized at $\lambda = 0.1$ as baseline values, step size ($\varepsilon = 0.01$) follows SVGD convergence theory, kernel bandwidth uses median heuristic, and projection counts scale with state dimensionality. This principled approach ensures reproducibility while minimizing domain-specific tuning.
    
    Notable domain-specific adjustments appear in the comparative methods, revealing insights into their computational strategies. DVRL's belief dimension scales proportionally with state complexity, using half the state dimension in Light-Dark (5) and Kidnapped Robot (10), but matching the full dimension in Target Tracking (20), suggesting a more expressive belief representation is needed for the complex multi-target environment. Similarly, POMCPOW employs deeper search depths ($d_{\text{max}}=5$) and more simulations ($n_{\text{sim}}=100$) with higher exploration constants ($c_{\text{expl}}=50.0$) in the challenging Kidnapped Robot environment compared to the other domains ($d_{\text{max}}=3$, $n_{\text{sim}}=50$, $c_{\text{expl}}=10.0$), indicating increased computational budget allocation for the perceptually ambiguous scenario with multiple similar landmarks. These systematic hyperparameter adjustments reflect a deliberate balance between maintaining algorithmic consistency and adapting to domain-specific challenges.

\section{Analysis of Baseline Methods} \label{sec:baseline_analysis}
    This section provides a detailed analysis of existing belief approximation methods and their fundamental limitations in addressing the challenges of high-dimensional, multi-modal belief distributions with complex correlation structures in POMDPs. We examine three categories of approaches: deterministic variational methods (SVGD), particle-based sampling methods (SIR filters~\citep{particle_filters_gordon} and their POMDP extensions like POMCPOW~\citep{pomcpow_sunberg}, POMCP~\citep{pomcp_silver}, ARDESPOT~\citep{ardespot_somani}), and parametric neural representations (DVRL~\citep{dvrl_igl}, DRQN~\citep{drqn_hausknecht}, ADRQN~\citep{adrqn_zhu}). While each category offers unique advantages---SVGD's deterministic particle evolution, particle filters' theoretical convergence guarantees, and neural methods' computational efficiency---we demonstrate how their core assumptions and algorithmic choices prevent them from simultaneously maintaining multi-modal coverage, preserving dimensional correlations, and scaling to high-dimensional belief spaces. Understanding these limitations not only motivates the design choices in ESCORT but also clarifies why a fundamentally new approach combining correlation-aware projections with temporal consistency constraints is necessary for accurate belief representation in complex POMDPs.
    
    Stein Variational Gradient Descent (SVGD)~\citep{svgd_liu} represents a significant advancement in Bayesian inference by providing a deterministic particle-based approach that bridges the gap between variational inference and sampling methods. The key insight of SVGD lies in its elegant formulation of particle updates through functional gradient descent in a reproducing kernel Hilbert space (RKHS), where particles evolve according to $\mathbf{x}_i^{t+1} = \mathbf{x}_i^t + \epsilon \phi^*(\mathbf{x}_i^t)$ with $\phi^*(\mathbf{x}) = \frac{1}{n}\sum_{j=1}^n [k(\mathbf{x}_j, \mathbf{x})\nabla_{\mathbf{x}_j} \log p(\mathbf{x}_j) + \nabla_{\mathbf{x}_j} k(\mathbf{x}_j, \mathbf{x})]$. This update mechanism ingeniously balances two forces: an attractive term $k(\mathbf{x}_j, \mathbf{x})\nabla_{\mathbf{x}_j} \log p(\mathbf{x}_j)$ that drives particles toward high-probability regions, and a repulsive term $\nabla_{\mathbf{x}_j} k(\mathbf{x}_j, \mathbf{x})$ that maintains particle diversity. While SVGD successfully addresses several limitations of traditional MCMC methods---particularly avoiding stochastic resampling and providing deterministic updates---it faces critical challenges when applied to high-dimensional POMDPs with complex belief distributions. The standard RBF kernel $k(\mathbf{x}, \mathbf{x}') = \exp(-\frac{1}{h}||\mathbf{x} - \mathbf{x}'||^2)$ suffers from kernel degeneracy as dimensionality increases, causing kernel values to become nearly uniform and weakening the essential repulsive forces needed to prevent mode collapse. Moreover, SVGD's isotropic kernel treats all dimensions uniformly, failing to capture the anisotropic nature of belief distributions where correlation strengths vary significantly across dimension pairs---a critical limitation when representing beliefs about interdependent state variables in realistic POMDP scenarios.
    
    Deep Variational Reinforcement Learning (DVRL) \cite{dvrl_igl} represents a sophisticated integration of variational inference with deep reinforcement learning, introducing an important inductive bias that enables agents to learn generative models of the environment and perform inference within those models. The key innovation of DVRL lies in its particle-based belief representation $\hat{b}_t = \{(h_t^k, z_t^k, w_t^k)\}_{k=1}^K$, where each particle consists of a deterministic RNN hidden state $h_t^k$, a stochastic latent variable $z_t^k$, and an importance weight $w_t^k$. This approach elegantly combines the expressiveness of neural networks with the flexibility of particle filters, using a variational autoencoder framework to jointly optimize an evidence lower bound (ELBO) alongside the reinforcement learning objective: $\mathcal{L}_t^{\text{DVRL}} = \mathcal{L}_t^A + \lambda_H \mathcal{L}_t^H + \lambda_V \mathcal{L}_t^V + \lambda_E \mathcal{L}_t^{\text{ELBO}}$. While DVRL successfully demonstrates that learning internal generative models improves performance over pure RNN-based approaches, it faces limitations when confronted with the specific challenges of high-dimensional belief spaces with complex correlation structures. The particle updates in DVRL follow standard importance sampling with resampling, where weights are computed as $w_t^k = \frac{p_\theta(z_t^k|h_{t-1}^{u_t^k}, a_{t-1})p_\theta(o_t|h_{t-1}^{u_t^k}, z_t^k, a_{t-1})}{q_\phi(z_t^k|h_{t-1}^{u_t^k}, a_{t-1}, o_t)}$, treating all state dimensions uniformly without mechanisms to capture or preserve dimensional dependencies. Furthermore, while the resampling step helps maintain particle diversity, it can disrupt correlation structures between state variables, and the lack of explicit temporal consistency constraints allows for potentially unrealistic belief transitions between timesteps---limitations that become particularly pronounced in environments where state variables exhibit strong interdependencies and beliefs must evolve smoothly over time.
    
    POMCPOW (Partially Observable Monte Carlo Planning with Observation Widening)~\citep{pomcpow_sunberg} extends POMCP to handle continuous observation spaces through double progressive widening (DPW) and weighted particle filtering. While standard POMCP suffers from belief collapse to single particles in continuous spaces—leading to QMDP-like policies that ignore information value—POMCPOW maintains weighted particle collections where each particle's weight is proportional to the observation likelihood $Z(o|s,a,s')$. This weighting mechanism prevents complete belief degeneracy and enables some information-gathering behavior. However, POMCPOW still faces critical limitations in representing complex belief distributions. First, it lacks explicit mechanisms to model correlation structures between state dimensions, treating particles independently without capturing the interdependencies crucial for realistic POMDPs. Second, the approach provides no temporal consistency constraints, allowing abrupt belief transitions that can destroy previously learned structures. Third, despite the weighting scheme, POMCPOW remains vulnerable to particle degeneracy in high-dimensional spaces where the effective sample size diminishes rapidly. These limitations mean that while POMCPOW improves upon basic POMCP for continuous observations, it cannot adequately represent the high-dimensional, multi-modal, correlated belief distributions that characterize complex POMDP domains.
    
    Beyond the methods discussed above, several other approaches have been proposed for belief approximation in POMDPs. ARDESPOT (Anytime Regularized DEterminized Sparse Partially Observable Tree)~\citep{ardespot_somani} uses determinized sparse sampling with regularization to scale POMCP to larger problems but still relies on unweighted particles that cannot capture complex correlation structures. AdaOPS (Adaptive Online Packing-guided Search)~\citep{adaops_wu} improves upon POMCP by adaptively selecting action and observation branches using packing constraints, though it remains limited by particle degeneracy in high-dimensional continuous spaces. DRQN (Deep Recurrent Q-Learning)~\citep{drqn_hausknecht} uses recurrent neural networks to compress observation histories into fixed-dimensional vectors but struggles to maintain distinct hypotheses for multi-modal beliefs. ADRQN~\citep{adrqn_zhu} augments DRQN with auxiliary tasks and attention mechanisms to better capture uncertainty, yet the fixed-size representation still cannot adapt to varying belief complexity. FORBES (Flow-based Recurrent Belief State Learning)~\citep{forbes_chen} employs normalizing flows to learn more expressive belief representations but requires extensive offline training and may not generalize well to novel scenarios. SBRL (Set-membership Belief State-based Reinforcement Learning)~\citep{sbrl_wei} maintains set-based belief representations that can capture some uncertainty structure but lacks the flexibility to model arbitrary multi-modal distributions with complex correlations. While each of these methods addresses specific aspects of belief representation, none provides a comprehensive solution for maintaining accurate, multi-modal beliefs with intricate correlation structures in high-dimensional continuous POMDPs.

\section{Domains} \label{sec:app_domains}
    \subsection{Light Dark 10D} \label{sec:light_dark_10d}     
        \begin{figure}[htbp]
            \centering
            \includegraphics[width=\textwidth]{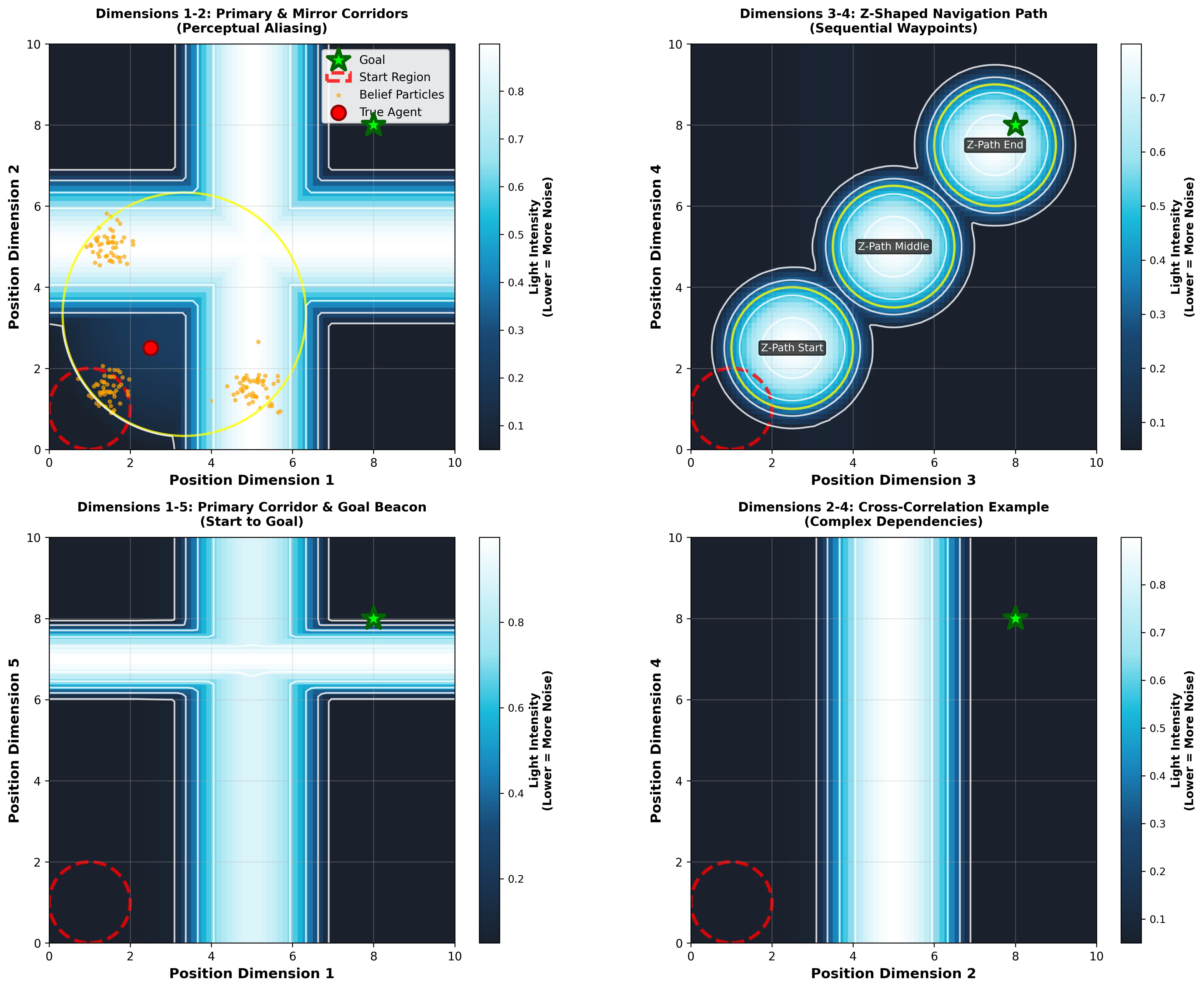}
            \caption{Light Dark 10D POMDP Environment. Four 2D projections of the 10D state space (5D position + 5D velocity) showing: \textbf{(Top-left)} Symmetric corridors creating perceptual aliasing with multi-modal belief particles (orange) splitting in dark regions; \textbf{(Top-right)} Z-shaped navigation path with sequential waypoints; \textbf{(Bottom-left)} Start-to-goal trajectory from dark region (red circle) to goal beacon (green star); \textbf{(Bottom-right)} Cross-dimensional correlations. Light intensity maps (blue-to-white) encode observation noise $\sigma^2(\mathbf{x}) = 0.5^2 \times (1 - L(\mathbf{x})) + 0.01^2$, with contour lines marking light boundaries. The environment challenges belief approximation through high dimensionality, multi-modality from symmetric patterns, and strong position-velocity correlations ($\rho = 0.8$).}
            \label{fig:lightdark_10d_domain}
        \end{figure}

        The Light-Dark Navigation environment extends the classical POMDP testbed to a high-dimensional setting that exhibits the fundamental challenges addressed by ESCORT. The environment operates in a 10-dimensional continuous state space $\mathcal{S} \subset \mathbb{R}^{10}$, decomposed into position coordinates $\mathbf{x} = (x_1, x_2, x_3, x_4, x_5) \in [0, 10]^5$ and velocity components $\mathbf{v} = (v_1, v_2, v_3, v_4, v_5) \in \mathbb{R}^5$. The agent can apply forces through 10 discrete actions $\mathcal{A} = \{0, 1, \ldots, 9\}$, where action $a = 2i$ applies positive force in dimension $i$ and $a = 2i + 1$ applies negative force. The agent receives noisy 5-dimensional position observations $\mathbf{o} \in \mathbb{R}^5$ with noise variance $\sigma^2(\mathbf{x}) = \sigma_{\text{base}}^2 \cdot (1 - L(\mathbf{x})) + \sigma_{\text{min}}^2$, where $L(\mathbf{x}) \in [0,1]$ represents the light intensity at position $\mathbf{x}$, $\sigma_{\text{base}} = 0.5$, and $\sigma_{\text{min}} = 0.01$. As illustrated in Figure~\ref{fig:lightdark_10d_domain}, the four 2D projections reveal the complex spatial structure, with light regions (blue-to-white gradient) providing precise observations and dark regions inducing high uncertainty.

        The environment contains seven strategically placed light regions that create complex belief landscapes and perceptual aliasing. The Primary Corridor (centered at $(5, 0, 0, 0, 0)$ with radius 2.0 and intensity 0.9) and Mirror Corridor (centered at $(0, 5, 0, 0, 0)$ with identical parameters) create symmetric patterns that induce multi-modal beliefs, as demonstrated by the orange belief particles in Figure~\ref{fig:lightdark_10d_domain} (top-left). Three connected Z-path segments in dimensions 3-4 provide sequential waypoints: Start at $(0, 0, 2.5, 2.5, 0)$, Middle at $(0, 0, 5.0, 5.0, 0)$, and End at $(0, 0, 7.5, 7.5, 0)$, each with radius 1.5 and intensity 0.8. A Goal Beacon near the target at $(0, 0, 0, 0, 7.0)$ provides high-precision observations (intensity 1.0), while an Ambiguous Region at $(3.33, 3.33, 3.33, 0, 0)$ with low intensity (0.4) further complicates belief maintenance. The light intensity function is computed as $L(\mathbf{x}) = \max(0.05, \max_j \{ I_j (1 - (d_j(\mathbf{x})/r_j)^2) \cdot \mathbb{I}[d_j(\mathbf{x}) < r_j] \})$, where $I_j$, $r_j$, and $d_j(\mathbf{x})$ are the intensity, radius, and distance to the $j$-th light region center. 

        The state evolution incorporates complex correlation structures that challenge standard belief approximation methods. The transition model follows $\mathbf{x}_{t+1} = \mathbf{x}_t + \Delta t \cdot \mathbf{v}_t + \boldsymbol{\eta}_{\mathbf{x}}$ and $\mathbf{v}_{t+1} = \mathbf{v}_t + \mathbf{f}_t - \gamma \mathbf{v}_t + \boldsymbol{\eta}_{\mathbf{v}}$, where $\mathbf{f}_t$ is the applied force (magnitude 0.1), $\gamma = 0.1$ is the damping coefficient, $\Delta t = 0.1$ is the time step, and $[\boldsymbol{\eta}_{\mathbf{x}}; \boldsymbol{\eta}_{\mathbf{v}}] \sim \mathcal{N}(\mathbf{0}, \boldsymbol{\Sigma}_{\text{corr}})$ represents correlated process noise. The correlation matrix $\boldsymbol{\Sigma}_{\text{corr}} \in \mathbb{R}^{10 \times 10}$ encodes strong position-velocity coupling ($\rho(x_i, v_i) = 0.8$), adjacent position correlations ($\rho(x_i, x_{i+1}) = 0.5$), adjacent velocity correlations ($\rho(v_i, v_{i+1}) = 0.6$), cross-dimensional dependencies ($\rho(x_1, x_3) = \rho(x_2, x_4) = 0.4$), and velocity interactions ($\rho(v_1, v_2) = 0.7$, $\rho(v_1, v_3) = 0.5$). Additionally, in very dark regions ($L(\mathbf{x}) < 0.1$), the observation model introduces dimensional identity confusion with probability 0.2, randomly swapping dimensions 1-2 or 3-4 to create further observational ambiguity.

        Performance evaluation focuses on navigation from a random starting position in the dark region $[0, 2]^5$ to the goal position $(8, 8, 8, 8, 8)$, with success defined as reaching within 0.5 units Euclidean distance. The primary metric is position error $\|\hat{\mathbf{x}} - \mathbf{x}_{\text{true}}\|_2$, where $\hat{\mathbf{x}}$ represents the belief mean estimate and $\mathbf{x}_{\text{true}}$ is the true agent position. The reward function combines navigation progress with a step penalty: $r_t = -0.1 \cdot \|\mathbf{x}_t - \mathbf{x}_{\text{goal}}\|_2 - 0.1$, encouraging both goal-directed behavior and efficient planning. This environment provides a rigorous testbed for evaluating ESCORT's ability to maintain accurate, multi-modal belief representations in high-dimensional spaces with complex correlation structures, directly addressing the fundamental challenges that motivate our approach.

    \subsection{Kidnapped Robot Problem} \label{sec:kidnapped_robot}    
        \begin{figure}[htbp]
            \centering
            \includegraphics[width=0.7\textwidth]{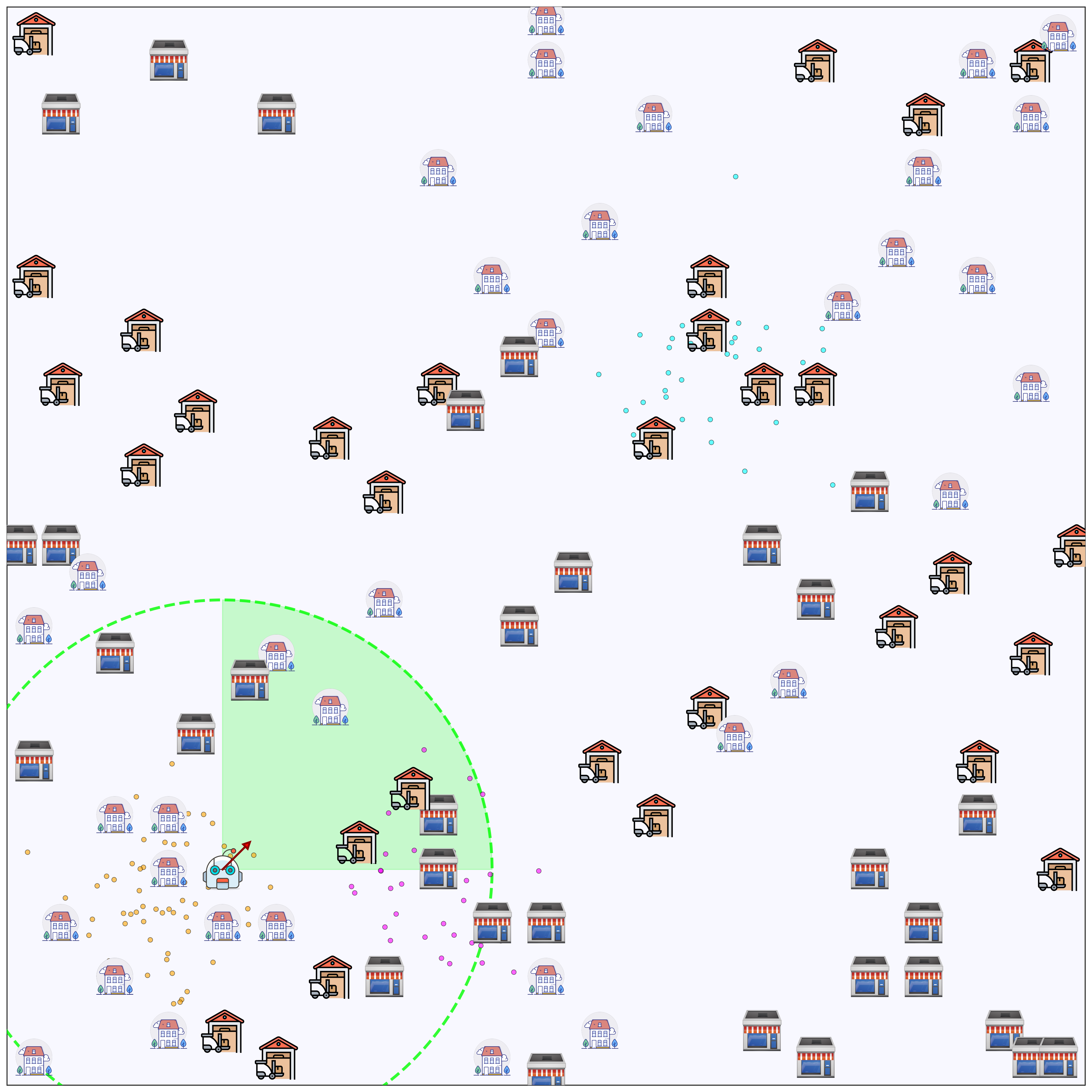}
            \caption{Kidnapped Robot Problem visualization showing domain. The robot icon (blue eyes with red antenna) indicates the true robot position and orientation. Various landmark types are distributed across the map: houses (red roofs), shops (red and white striped awnings), and warehouses (with forklift symbols), creating perceptually similar patterns that cause aliasing. The green dashed circle shows the sensor range (radius = 5), and green dashed lines indicate the 90° field of view. Orange dots represent belief particles from ESCORT, with particle density shown as a yellow-red heatmap overlay. The robot must localize itself despite ambiguous observations from these visually similar landmark configurations.}
            \label{fig:kidnapped_robot}
        \end{figure}

        The Kidnapped Robot Problem (present in Figure~\ref{fig:kidnapped_robot}) represents a classical robotics localization challenge scaled to high dimensionality with complex correlation structures. The robot operates within a 20×20 map containing various landmarks of different types—houses, shops, and warehouses—arranged in perceptually similar patterns that create fundamental ambiguity in observations. The state space is 20-dimensional, comprising 2D position $(x, y) \in [0, 20]^2$, orientation $\theta \in [0, 2\pi)$, velocity and steering parameters $(v, s) \in \mathbb{R}^2$, sensor calibration parameters $\mathbf{c} \in \mathbb{R}^5$, and environmental feature descriptors $\mathbf{f} \in \mathbb{R}^{10}$ with $\|\mathbf{f}\|_2 = 1$. The robot's sensor has a limited range of 5 units and a 90° field of view, generating observations consisting of distance measurements and feature similarity scores for visible landmarks.

        The environment incorporates strong correlation structures that reflect realistic robotic systems. Position and orientation exhibit correlation coefficients of 0.6, while position-velocity correlations reach 0.8, representing coupled dynamics typical in mobile robotics. Sensor calibration parameters maintain internal correlations of 0.4 and influence feature descriptors with coefficients of 0.3, modeling sensor drift and environmental perception coupling. The transition dynamics follow the update equations: $x_{t+1} = x_t + v_t \cos(\theta_t) \Delta t$, $y_{t+1} = y_t + v_t \sin(\theta_t) \Delta t$, and $\theta_{t+1} = \theta_t + \Delta\theta_{\text{action}}$, where actions modify orientation ($\Delta\theta = \pm 0.1$) or maintain position. Correlated noise is applied using the full correlation matrix $\mathbf{C} \in \mathbb{R}^{20 \times 20}$, generating realistic multi-variate updates that preserve dimensional dependencies.
        
        The fundamental challenge arises from perceptual aliasing where multiple landmark configurations produce nearly identical observations, creating multi-modal belief distributions. The map contains repeating patterns such as clusters of houses, shops, and warehouses at different locations that generate similar feature vectors with small perturbations ($\sigma = 0.1$). When the robot observes these patterns, the belief distribution develops multiple modes corresponding to each plausible location, with correlation structures linking position hypotheses to consistent orientation and velocity estimates. This multi-modality, combined with the high-dimensional correlated state space, creates the precise challenge that ESCORT addresses through its correlation-aware particle evolution and temporal consistency constraints.
        
        Performance evaluation focuses on localization accuracy measured as position error $\epsilon_{\text{pos}} = \|\mathbf{p}_{\text{true}} - \mathbb{E}[\mathbf{p}_{\text{belief}}]\|_2$, where $\mathbf{p}_{\text{true}} \in \mathbb{R}^2$ represents the true robot position and $\mathbb{E}[\mathbf{p}_{\text{belief}}]$ denotes the expected position from the belief distribution. The reward structure implements $r_t = -1$ per timestep to encourage efficient localization, with episode termination based on convergence criteria or maximum step limits, ensuring that methods must balance exploration of multiple hypotheses with rapid convergence to the true robot location.

    \subsection{Multiple Target Tracking} \label{sec:target_tracking}
        \begin{figure}[t]
            \centering
            \includegraphics[width=0.7\textwidth]{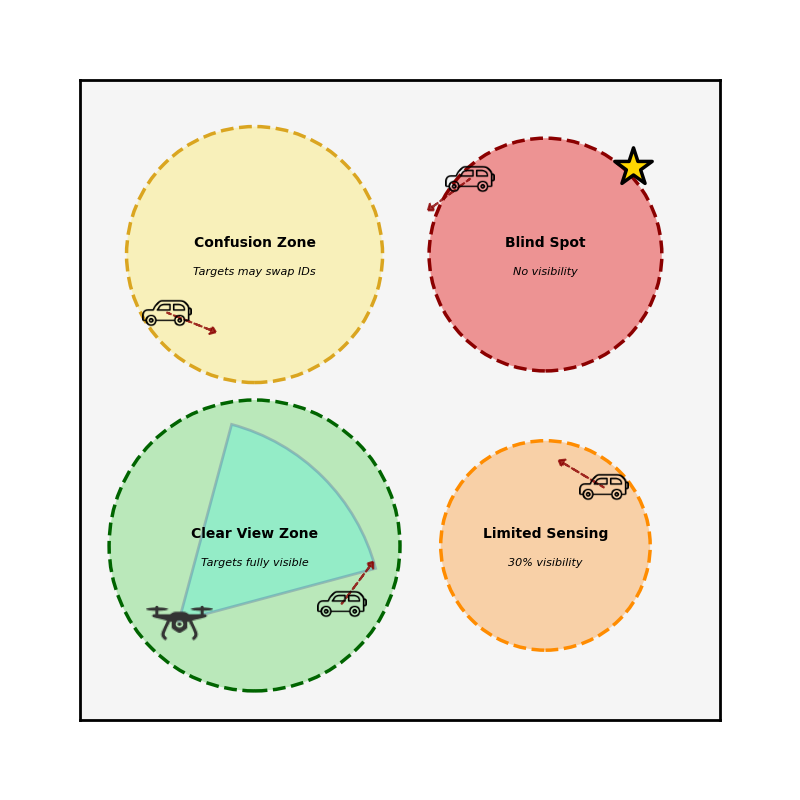}
            \caption{Multiple Target Tracking Environment. An agent (drone) with limited field of view (cyan wedge) must navigate to the goal (star) while tracking multiple independently moving targets (cars with velocity arrows) across zones with varying observability: Clear View (full visibility), Limited Sensing (30\% visibility), Confusion Zone (identity swaps possible), and Blind Spot (no visibility). The 20D continuous state space and partial observability create multi-modal belief distributions.}
            \label{fig:mtt_environment}
        \end{figure}

        The Multiple Target Tracking domain extends classical pursuit-evasion scenarios to test belief approximation under high dimensionality, multi-modality, and complex correlations. An agent must navigate to a goal position while maintaining awareness of four independently moving targets despite varying observability conditions that create ambiguous, multi-modal belief distributions.

        The environment consists of a continuous $10 \times 10$ space with state $\mathbf{s} \in \mathbb{R}^{20}$ comprising the agent's position and velocity $\mathbf{s}_a = [x_a, y_a, v_{x_a}, v_{y_a}]^T$ and four targets' states $\mathbf{s}_{t_i} = [x_{t_i}, y_{t_i}, v_{x_{t_i}}, v_{y_{t_i}}]^T$ for $i \in \{1,2,3,4\}$. The agent receives partial observations $\mathbf{o} \in \mathbb{R}^{10}$ containing noisy position measurements of itself and targets: $\mathbf{o} = [x_a + \epsilon_a, y_a + \epsilon_a, x_{t_1} + \epsilon_{t_1}, y_{t_1} + \epsilon_{t_1}, \ldots]^T$, where observation noise $\epsilon$ varies based on spatial zones. As illustrated in Figure~\ref{fig:mtt_environment}, four distinct visibility zones create varying observation conditions: \textit{Clear View} (green zone, visibility $\nu = 1.0$, full observations), \textit{Limited Sensing} (orange zone, $\nu = 0.3$, high noise), \textit{Confusion Zone} (yellow zone, $\nu = 0.7$ but 30\% chance of identity swaps between targets), and \textit{Blind Spot} (red zone, $\nu = 0.0$, no target observations).

        The system dynamics exhibit strong correlations through physical constraints and environmental influences. State transitions follow $\mathbf{s}_{t+1} = f(\mathbf{s}_t, a_t) + \boldsymbol{\eta}_t$, where $f$ incorporates velocity damping ($\lambda = 0.1$), agent acceleration from discrete actions $a_t \in \{+x, -x, +y, -y\}$, and environmental flow fields $\mathbf{F}(\mathbf{x})$ that create correlated target movements. The correlation matrix $\mathbf{C} \in \mathbb{R}^{20 \times 20}$ captures position-velocity couplings within entities ($C_{x,v_x} = C_{y,v_y} = 0.8$) and inter-target dependencies from flocking behavior ($C_{t_i,t_j} = 0.4$ for positions, $0.6$ for velocities). The red arrows in Figure~\ref{fig:mtt_environment} indicate the current velocity directions of each target, which are influenced by both individual dynamics and collective flow patterns. Collision avoidance introduces additional correlations through repulsive forces when $\|\mathbf{x}_{t_i} - \mathbf{x}_{t_j}\| < 1.0$.

        The combination of limited sensing and zone-dependent observability creates severe challenges for belief representation. When targets enter the Blind Spot (as shown for one target in Figure~\ref{fig:mtt_environment}), the belief must maintain hypotheses about their possible locations, creating multi-modal distributions. The Confusion Zone induces additional modes when identity swaps occur—if the agent observes a target at position $\mathbf{x}_{obs}$, the belief must consider it could be any of the targets whose last known positions were nearby. This ambiguity compounds over time as $P(o_t | s_t) = \prod_{i=1}^{4} P(o_{t,i} | s_{t,i}, \text{zone}(s_{t,i}))$, where zone-dependent likelihoods create sharp discontinuities. The agent's limited field of view (60° cone shown in cyan in Figure~\ref{fig:mtt_environment}) further exacerbates partial observability, as targets outside the FOV receive no updates regardless of zone visibility.

        The reward function balances navigation and safety objectives: $r_t = -\alpha \|\mathbf{x}_a - \mathbf{x}_{goal}\| - \beta - \sum_{i=1}^{4} \mathbb{1}[\|\mathbf{x}_a - \mathbf{x}_{t_i}\| < \delta]$, where $\alpha = 0.1$ weights distance to goal (marked by the star in Figure~\ref{fig:mtt_environment}), $\beta = 0.05$ provides step penalty, and collision penalty is triggered when agent-target distance falls below $\delta = 0.5$. Episode success requires reaching $\|\mathbf{x}_a - \mathbf{x}_{goal}\| < 0.5$. Performance is evaluated by the mean position error between true and estimated agent position across belief particles: $\text{error} = \|\mathbf{x}_a^{\text{true}} - \mathbb{E}_{\mathbf{b}}[\mathbf{x}_a]\|$, where the belief mean is computed from particle representation.

    \subsection{Visual Observation Environments: Flickering Atari}
        To evaluate ESCORT's effectiveness with high-dimensional visual observations under severe partial observability, we conducted experiments on Flickering Atari environments~\citep{gymnasium_towers,atari_bellemare,atari_machado}. These environments use a flickering mechanism with 50\% probability of blank screen observations and single-frame inputs~\citep{dvrl_igl,dpfrl_ma}, creating substantial uncertainty about the current state. We compare against DVRL on four standard Atari games with different complexity characteristics, following the experimental protocol from the DVRL paper for fair comparison. Table~\ref{tab:atari_results} presents the performance results.

        \begin{table}[htbp]
        \centering
        \caption{Performance on Flickering Atari Environments (Higher is Better)}
        \label{tab:atari_results}
        \begin{tabular}{lcc}
        \toprule
        \textbf{Environment} & \textbf{ESCORT} & \textbf{DVRL} \\
        \midrule
        Pong & $17.97 \pm 3.74$ & $\mathbf{18.17 \pm 2.67}$ \\
        IceHockey & $\mathbf{-4.63 \pm 0.19}$ & $-4.88 \pm 0.17$ \\
        MsPacman & $\mathbf{3179.7 \pm 356.7}$ & $2221 \pm 199$ \\
        Asteroids & $\mathbf{1787.7 \pm 239.6}$ & $1539 \pm 73$ \\
        \bottomrule
        \end{tabular}
        \end{table}
        
        The results demonstrate ESCORT's effectiveness with raw visual observations under severe partial observability. In simple reactive environments (Pong, IceHockey), ESCORT achieves comparable performance to DVRL despite temporal consistency potentially over-constraining fast reactive dynamics. However, ESCORT significantly outperforms DVRL in complex multi-object tracking environments (MsPacman, Asteroids) where multiple ghosts/asteroids create multi-modal beliefs with crucial position-velocity correlations. ESCORT's deterministic particle evolution maintains these multiple hypotheses and dimensional dependencies effectively, while DVRL's VAE compression struggles to preserve the multi-modal structure necessary for accurate tracking under flickering observations.

\section{Synthetic multi-modal distributions} \label{sec:synthetic}
    To systematically evaluate ESCORT's capability in addressing the fundamental challenges of belief representation—high dimensionality, multi-modality, and complex correlation structures—we designed a comprehensive suite of synthetic benchmark distributions. These controlled experiments allow us to isolate and measure specific aspects of belief approximation performance that are difficult to disentangle in real POMDP environments. By progressively increasing dimensionality from 1D to 20D while maintaining consistent multi-modal characteristics, we can observe how each method's performance degrades with the curse of dimensionality and assess their ability to preserve critical distributional properties such as mode coverage and correlation structures. This systematic evaluation complements our POMDP experiments by providing precise quantitative metrics for belief representation fidelity.

    \subsection{Evaluation Metrics}\label{subsec:metrics_computation}
        To quantitatively assess the quality of belief approximation across different dimensionalities, we employ a comprehensive set of metrics that capture complementary aspects of distributional fidelity (results presented in Table~\ref{tab:belief_assessment}):
        
        \textbf{Maximum Mean Discrepancy (MMD)} measures the distance between two distributions in a reproducing kernel Hilbert space. For samples $\{x_i\}_{i=1}^n \sim p$ and $\{y_j\}_{j=1}^m \sim q$, the empirical MMD with RBF kernel $k(x,y) = \exp(-\gamma||x-y||^2)$ is computed as:
        \begin{equation}
            \text{MMD}^2(p,q) = \frac{1}{n^2}\sum_{i,j=1}^{n} k(x_i,x_j) + \frac{1}{m^2}\sum_{i,j=1}^{m} k(y_i,y_j) - \frac{2}{nm}\sum_{i=1}^{n}\sum_{j=1}^{m} k(x_i,y_j)
        \end{equation}
        This metric captures overall distributional similarity, with lower values indicating better approximation quality. The kernel parameter $\gamma = 0.5$ provides sensitivity to both local and global distribution differences.
        
        \textbf{Wasserstein Distance} (1-Wasserstein or Earth Mover's Distance) quantifies the minimum ``cost'' of transforming one distribution into another:
        \begin{equation}
            W_1(p,q) = \inf_{\gamma \in \Gamma(p,q)} \int ||x-y||_1 \, d\gamma(x,y)
        \end{equation}
        where $\Gamma(p,q)$ denotes the set of all joint distributions with marginals $p$ and $q$. For 1D distributions, this reduces to the $L_1$ distance between inverse cumulative distribution functions, efficiently computed via sorting. Unlike MMD, Wasserstein distance explicitly accounts for the metric structure of the space, making it particularly sensitive to mode locations.
        
        \textbf{Sliced Wasserstein Distance} extends Wasserstein distance to high dimensions by projecting distributions onto one-dimensional subspaces:
        \begin{equation}
            \text{SW}_1(p,q) = \int_{\mathbb{S}^{d-1}} W_1(\mathcal{R}_\theta p, \mathcal{R}_\theta q) \, d\sigma(\theta)
        \end{equation}
        where $\mathcal{R}_\theta$ denotes the Radon transform (projection) along direction $\theta \in \mathbb{S}^{d-1}$. This approach maintains computational efficiency while preserving geometric properties, computed via Monte Carlo approximation over random projections.
        
        \textbf{Mode Coverage Ratio} specifically evaluates multi-modal representation quality. Given target mode locations $\{\mu_k\}_{k=1}^K$ and approximating samples $\{x_i\}_{i=1}^n$, a mode $k$ is considered ``covered'' if:
        \begin{equation}
            \frac{|\{x_i : ||x_i - \mu_k||_2 < \tau\}|}{n} > 0.05 \cdot \frac{1}{K}
        \end{equation}
        where $\tau = 1.0$ is the coverage threshold. The metric returns the fraction of modes satisfying this criterion, directly measuring whether methods maintain all hypotheses or suffer from mode collapse.
        
        \textbf{Correlation Error} (for dimensions $\geq 2$) measures how well methods preserve inter-dimensional dependencies. Given true correlation matrix $C_{\text{true}}$ and approximated correlation matrix $C_{\text{approx}}$ computed from samples:
        \begin{equation}
            \text{Correlation Error} = ||C_{\text{true}} - C_{\text{approx}}||_F
        \end{equation}
        where $||\cdot||_F$ denotes the Frobenius norm. This metric is crucial for evaluating ESCORT's correlation-aware regularization mechanism, as preserving dimensional dependencies is essential for accurate belief representation in POMDPs.
        
        These metrics collectively provide a comprehensive evaluation framework: MMD and Wasserstein/Sliced Wasserstein capture global distributional fidelity, Mode Coverage Ratio explicitly quantifies multi-modal representation capability, and Correlation Error measures the preservation of dimensional dependencies critical for complex belief structures.

    \subsection{1D Multi-modal Gaussian Mixture Model} \label{subsec:1d_gmm} 
        \begin{figure}[htbp]
            \centering
            \includegraphics[width=0.8\textwidth]{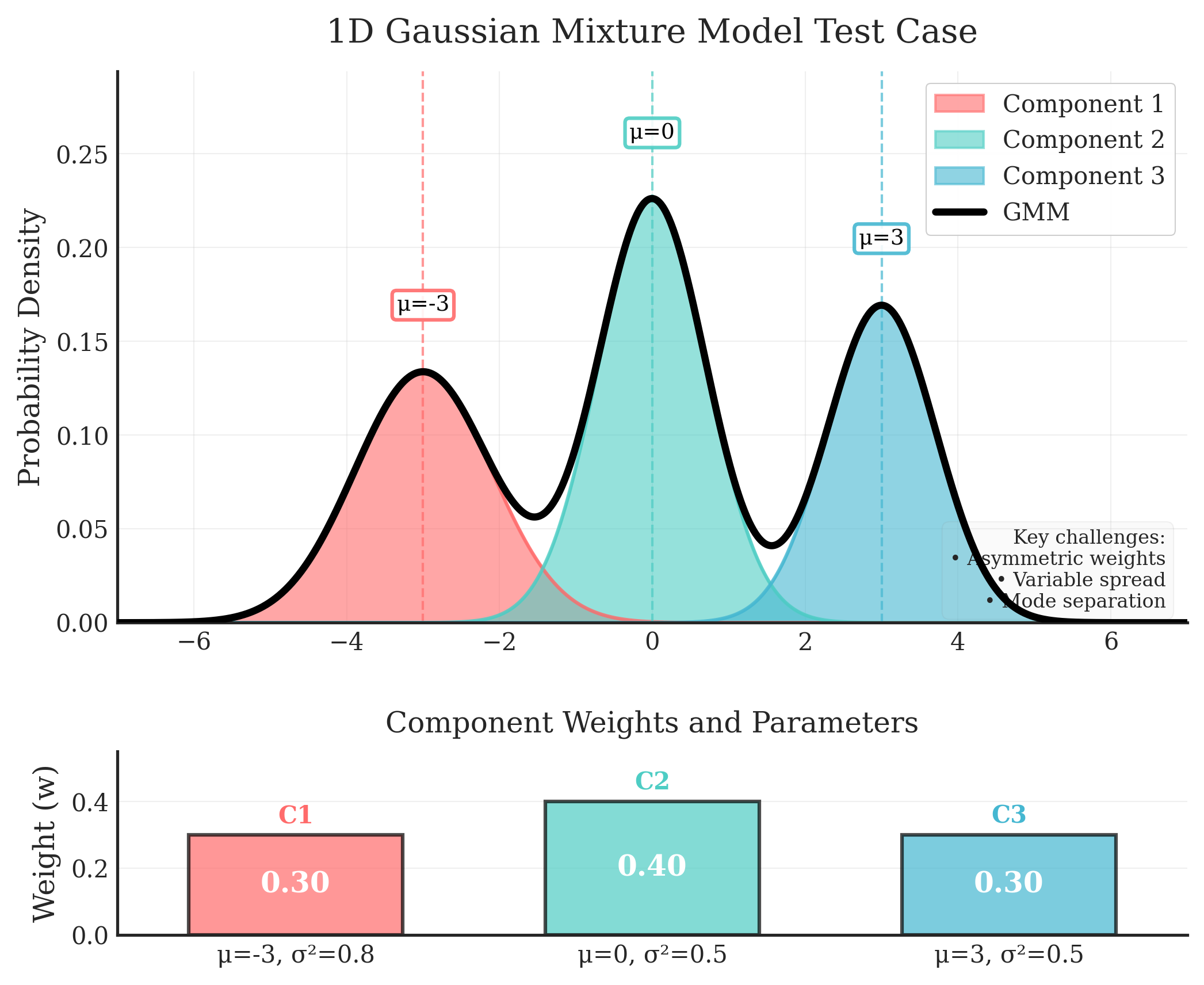}
            \caption{Visualization of the 1D Gaussian Mixture Model test case. The top panel shows the overall GMM density (black line) decomposed into three weighted components with different means and variances. Vertical dashed lines indicate component means, with annotations showing the precise locations. The bottom panel displays the component weights and variances, highlighting the asymmetric nature of the distribution that challenges belief approximation methods.}
            \label{fig:1d_gmm_test}
        \end{figure}
        
        Our 1D test case consists of a carefully designed Gaussian Mixture Model (GMM) that encapsulates the multi-modality challenge in its simplest form while remaining non-trivial for approximation methods. The target distribution is defined as:
        \begin{equation}
            p(x) = \sum_{k=1}^{3} w_k \mathcal{N}(x; \mu_k, \sigma_k^2)
        \end{equation}
        where the component parameters are $\mu_1 = -3.0$, $\mu_2 = 0.0$, $\mu_3 = 3.0$ for the means, $\sigma_1^2 = 0.8$, $\sigma_2^2 = 0.5$, $\sigma_3^2 = 0.5$ for the variances, and $w_1 = 0.3$, $w_2 = 0.4$, $w_3 = 0.3$ for the mixture weights.
        
        This configuration presents several challenges that mirror those encountered in POMDP belief representation. First, the unequal weights create an asymmetric distribution where methods must balance between accurately representing the dominant central mode ($w_2 = 0.4$) while maintaining sufficient particles at the less probable side modes. Second, the different variances, with the first component having larger spread ($\sigma_1^2 = 0.8$), test whether methods can adapt their particle density to match the local uncertainty structure. Third, the well-separated modes (6 units apart) ensure that methods cannot rely on a single concentrated particle cloud but must actively maintain multiple distinct hypotheses.
        
        As shown in Figure~\ref{fig:1d_gmm_test}, the resulting distribution exhibits clear separation between modes while maintaining smooth probability gradients that allow gradient-based methods like SVGD and ESCORT to navigate the landscape effectively. The asymmetric weights and variances create a more realistic test case than uniform mixtures, as real POMDP beliefs often exhibit similar heterogeneity due to varying observation quality across the state space. This 1D case serves as the foundation for understanding each method's behavior before examining how their performance scales to higher dimensions where additional challenges of correlation preservation and exponential volume growth emerge.

    \subsection{2D Correlated Gaussian Mixture Model} \label{subsec:2d_correlated_gmm}
        \begin{figure}[htbp]
            \centering
            \includegraphics[width=0.75\textwidth]{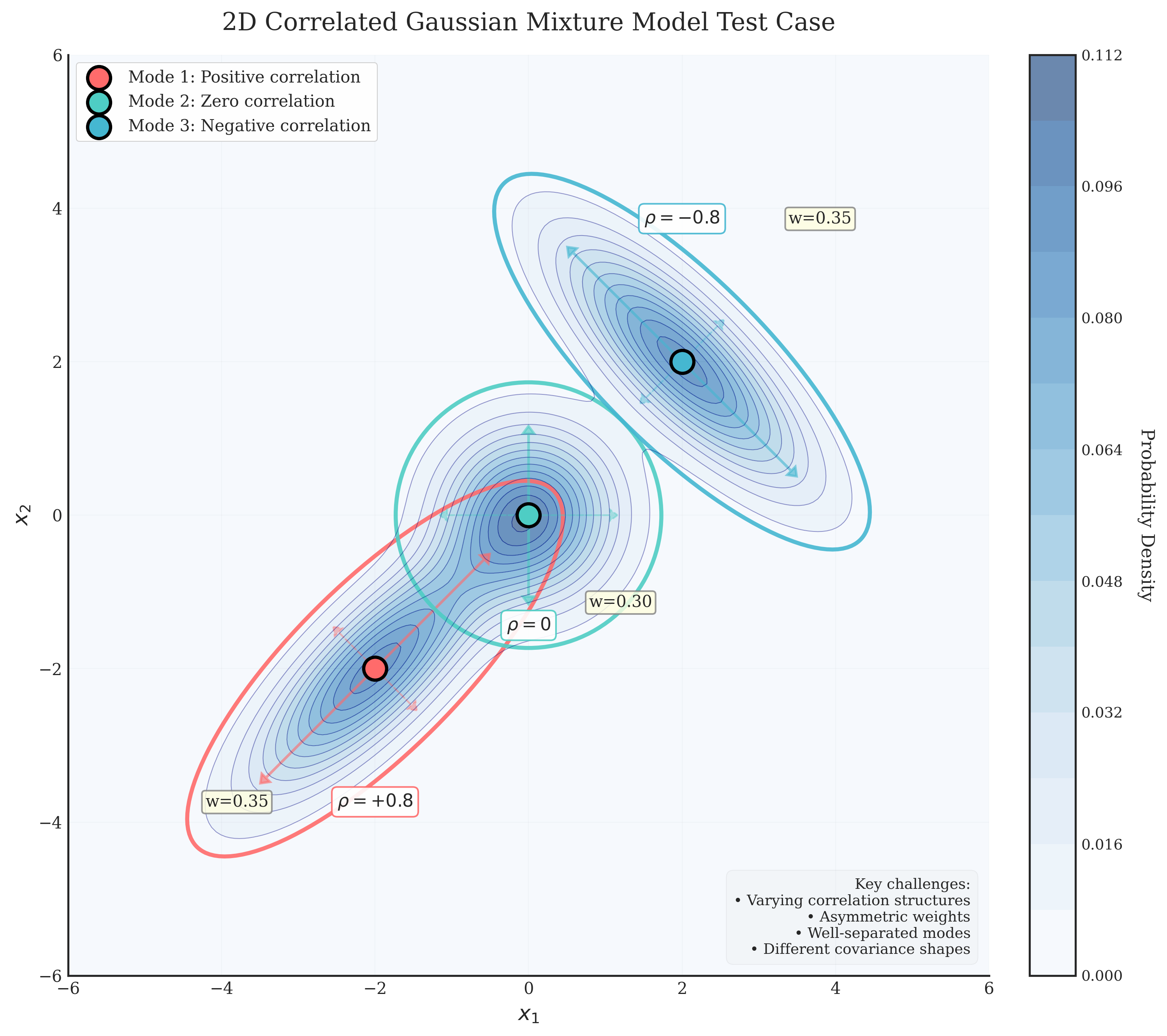}
            \caption{Visualization of the 2D Correlated Gaussian Mixture Model test case. The filled contours represent the overall probability density, while the three components are shown with their 95\% confidence ellipses. Arrows indicate the principal axes of each covariance matrix, illustrating the positive correlation (Mode 1, bottom-left), zero correlation (Mode 2, center), and negative correlation (Mode 3, top-right). The correlation coefficients $\rho$ and mixture weights $w_k$ are annotated for each component.}
            \label{fig:2d_correlated_gmm_test}
        \end{figure}

        Building upon the 1D evaluation, our 2D test case introduces the critical challenge of correlation structures between state dimensions. The target distribution is a three-component GMM designed to test each method's ability to preserve diverse correlation patterns:
        \begin{equation}
            p(\mathbf{x}) = \sum_{k=1}^{3} w_k \mathcal{N}(\mathbf{x}; \boldsymbol{\mu}_k, \boldsymbol{\Sigma}_k)
        \end{equation}
        where $\mathbf{x} = [x_1, x_2]^T$ and the component parameters are carefully chosen to create distinct correlation challenges.

        The three modes are positioned at $\boldsymbol{\mu}_1 = [-2, -2]^T$, $\boldsymbol{\mu}_2 = [0, 0]^T$, and $\boldsymbol{\mu}_3 = [2, 2]^T$ with weights $w_1 = 0.35$, $w_2 = 0.30$, and $w_3 = 0.35$. The critical distinguishing feature lies in their covariance structures:
        \begin{align}
            \boldsymbol{\Sigma}_1 &= \begin{bmatrix} 1.0 & 0.8 \\ 0.8 & 1.0 \end{bmatrix} \quad \text{(positive correlation, } \rho = 0.8\text{)} \\
            \boldsymbol{\Sigma}_2 &= \begin{bmatrix} 0.5 & 0.0 \\ 0.0 & 0.5 \end{bmatrix} \quad \text{(no correlation, } \rho = 0\text{)} \\
            \boldsymbol{\Sigma}_3 &= \begin{bmatrix} 1.0 & -0.8 \\ -0.8 & 1.0 \end{bmatrix} \quad \text{(negative correlation, } \rho = -0.8\text{)}
        \end{align}
        
        This configuration presents several interrelated challenges that directly test ESCORT's correlation-aware mechanisms. First, the varying correlation structures—from strong positive through zero to strong negative correlation—require methods to adapt their particle dynamics to match the local covariance geometry rather than applying uniform, isotropic updates. Second, the asymmetric weights create a subtle imbalance where methods must allocate appropriate computational resources to each mode while respecting their different shapes. Third, the well-separated modes (distance of $4\sqrt{2}$ units between adjacent modes) ensure that simple diffusion-based approaches cannot bridge the modes without explicit multi-modal handling.

        As illustrated in Figure~\ref{fig:2d_correlated_gmm_test}, the distribution creates a challenging landscape where each mode requires different treatment. The elliptical contours reveal how correlation structures fundamentally alter the shape of uncertainty regions: Mode 1's positive correlation creates an elongated ellipse along the diagonal, Mode 2's spherical shape reflects independent dimensions, while Mode 3's negative correlation produces an ellipse oriented perpendicular to the diagonal. These geometric differences are not merely aesthetic—they represent fundamentally different relationships between state variables that must be preserved during belief updates.

        The 2D case serves as a critical bridge between the simplicity of 1D and the complexity of high-dimensional spaces. While maintaining computational tractability for detailed analysis, it introduces the essential challenge of correlation preservation that becomes increasingly important in higher dimensions. Methods that fail to account for these correlation structures will either oversample along incorrect directions (wasting particles) or undersample critical regions (missing important probability mass), leading to poor belief approximation and suboptimal decision-making in POMDP applications.

    \subsubsection{3D Correlated Gaussian Mixture Model} \label{subsec:3d_correlated_gmm}
        \begin{figure}[htbp]
            \centering
            \includegraphics[width=0.48\textwidth]{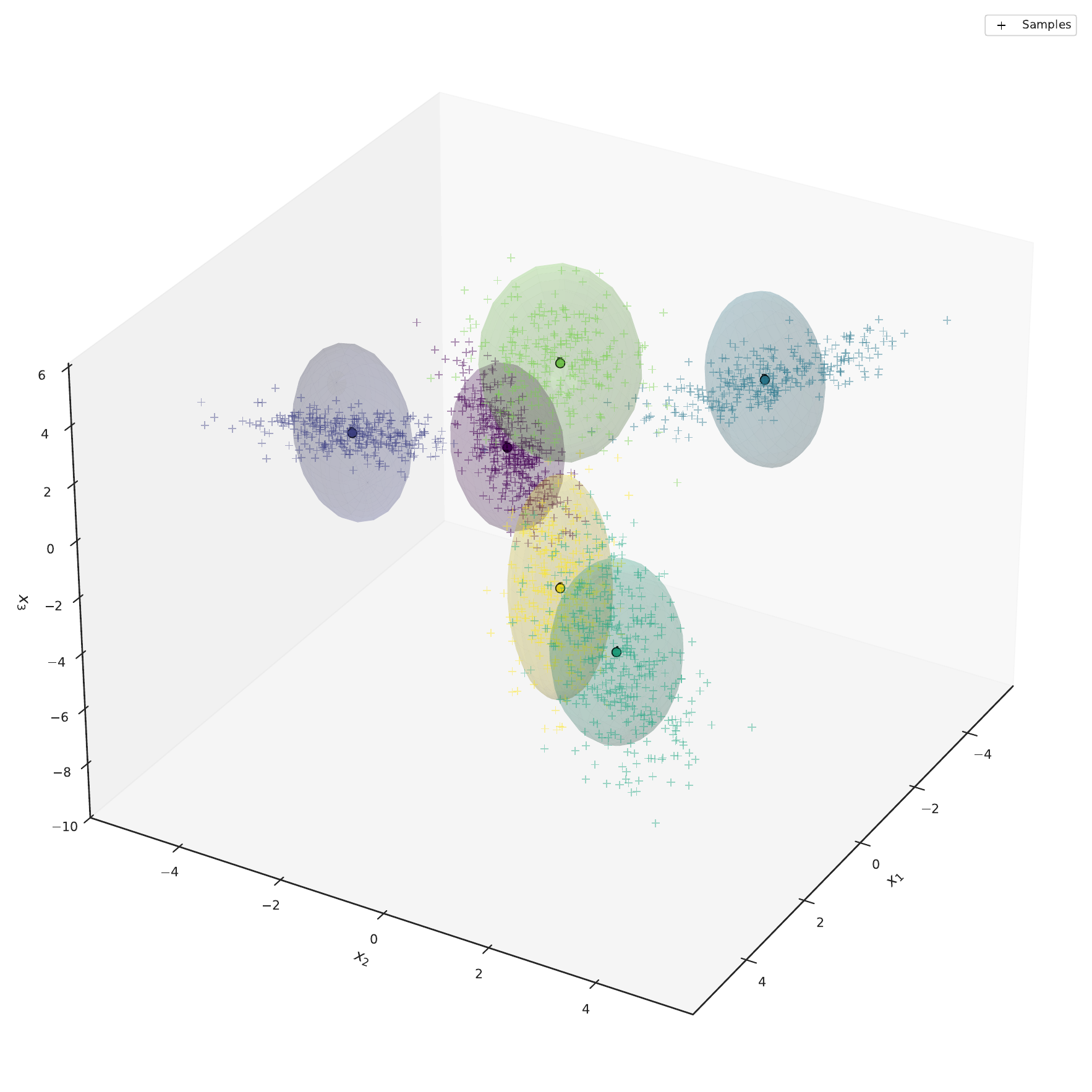}
            \includegraphics[width=0.48\textwidth]{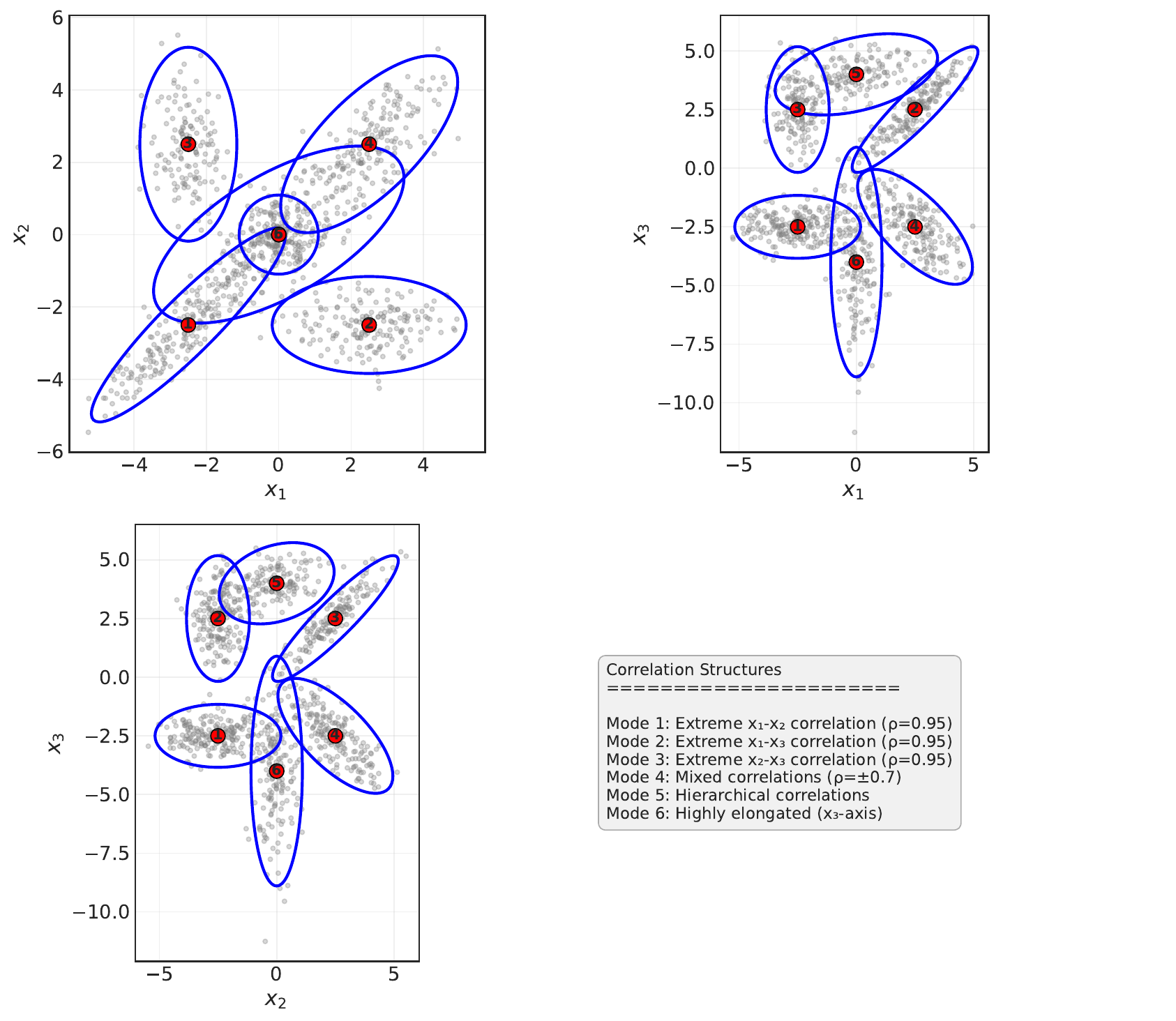}
            \caption{Three-dimensional Gaussian Mixture Model test distribution with complex correlation structures. \textbf{Left}: 3D visualization showing six modes with distinct correlation patterns, where ellipsoids represent 95\% confidence regions and colors indicate component membership. Mode 6's extreme elongation along the Z-axis and the planar correlations of Modes 1-3 are clearly visible. \textbf{Right}: Three 2D projections onto coordinate planes (XY, XZ, YZ) reveal the correlation patterns with a summary of correlation structures. The XY projection shows Mode 1's strong correlation; XZ projection highlights Mode 2's correlation; YZ projection displays Mode 3's structure. The correlation matrix for each mode determines the ellipsoid orientation and eccentricity.}
            \label{fig:3d_gmm}
        \end{figure}

        Building upon the correlation preservation challenges introduced in the 2D case, our 3D test extends the evaluation to capture the full complexity of belief distributions encountered in realistic POMDP scenarios. The target distribution is a six-component GMM specifically designed to test ESCORT's ability to preserve diverse three-dimensional correlation structures:

        \begin{equation}
        p(\mathbf{x}) = \sum_{k=1}^{6} w_k \mathcal{N}(\mathbf{x}; \boldsymbol{\mu}_k, \boldsymbol{\Sigma}_k)
        \end{equation}
        
        where $\mathbf{x} = [x_1, x_2, x_3]^T$ and the component parameters create distinct correlation challenges that directly correspond to belief structures in 3D state spaces.
        
        The six modes are positioned at $\boldsymbol{\mu}_1 = [-2.5, -2.5, -2.5]^T$, $\boldsymbol{\mu}_2 = [2.5, -2.5, 2.5]^T$, $\boldsymbol{\mu}_3 = [-2.5, 2.5, 2.5]^T$, $\boldsymbol{\mu}_4 = [2.5, 2.5, -2.5]^T$, $\boldsymbol{\mu}_5 = [0, 0, 4]^T$, and $\boldsymbol{\mu}_6 = [0, 0, -4]^T$ with weights $\mathbf{w} = [0.2, 0.15, 0.15, 0.2, 0.15, 0.15]^T$. Each mode exhibits a unique correlation pattern designed to challenge specific aspects of belief approximation:
        
        \begin{itemize}
            \item \textbf{Mode 1}: Extreme $x_1$-$x_2$ plane correlation ($\rho_{12} = 0.95$) with minimal $x_3$ variance, representing beliefs where two state variables are tightly coupled while the third remains independent—common in robotic systems where position coordinates are correlated but orientation varies freely.
            
            \item \textbf{Mode 2}: Extreme $x_1$-$x_3$ plane correlation ($\rho_{13} = 0.95$) with minimal $x_2$ variance, testing the ability to capture correlations across non-adjacent dimensions.
            
            \item \textbf{Mode 3}: Extreme $x_2$-$x_3$ plane correlation ($\rho_{23} = 0.95$) with minimal $x_1$ variance, completing the set of planar correlations.
            
            \item \textbf{Mode 4}: Complex mixed correlations with both positive ($\rho = 0.7$) and negative ($\rho = -0.7$) dependencies:
            \begin{equation}
            \boldsymbol{\Sigma}_4 = \begin{bmatrix}
            1.0 & 0.7 & -0.7 \\
            0.7 & 1.0 & -0.7 \\
            -0.7 & -0.7 & 1.0
            \end{bmatrix}
            \end{equation}
            This mode represents belief states where increasing confidence in one dimension simultaneously increases confidence in another while decreasing it in the third—a pattern observed in constrained optimization problems.
            
            \item \textbf{Mode 5}: Hierarchical correlations with varying magnitudes, where the correlation strength decreases with dimensional distance, modeling cascading uncertainty propagation in sequential state estimation.
            
            \item \textbf{Mode 6}: Highly elongated distribution along the $x_3$-axis ($\sigma_3^2 = 4.0$ while $\sigma_1^2 = \sigma_2^2 = 0.2$), testing the ability to maintain particles in extremely anisotropic distributions without collapse.
        \end{itemize}

        This 3D configuration presents compounded challenges beyond the 2D case. The curse of dimensionality begins to manifest more severely—while maintaining coverage of six modes in 3D requires only $6^{1/3} \approx 1.8\times$ more particles per dimension than in 2D, the variety of correlation patterns demands sophisticated particle dynamics. Methods must simultaneously: (1) maintain sufficient particles in each mode despite the increased volume, (2) preserve three distinct types of planar correlations, (3) handle mixed positive-negative correlations that create saddle-shaped uncertainty regions, and (4) prevent particle collapse in the highly elongated Mode 6.

        As shown in Figure~\ref{fig:3d_gmm}, the distribution creates a challenging landscape where each mode requires fundamentally different treatment. The planar correlations in Modes 1-3 require particles to align along specific 2D subspaces within the 3D space, while Mode 4's mixed correlations create a complex saddle structure that naive isotropic updates cannot capture. Mode 5's hierarchical structure tests whether methods can model correlations of varying strength, while Mode 6's extreme elongation along the $x_3$-axis challenges particle filters that typically assume roughly isotropic uncertainty.
        
        The results in Table~\ref{tab:belief_assessment} for the 3D experiment reveal ESCORT's advantages in this intermediate-dimensional space. While all methods maintain perfect mode coverage (1.0), indicating sufficient exploration capabilities in 3D, the correlation error metric reveals significant performance differences: ESCORT achieves 0.761 correlation error compared to SVGD's 0.819, DVRL's 0.882, and SIR's 1.003. This 7\% improvement over SVGD and 24\% over SIR demonstrates that ESCORT's correlation-aware projections effectively preserve the complex interdependencies between state variables. The MMD and Sliced Wasserstein metrics further confirm ESCORT's superior distributional approximation, with particularly strong performance in capturing the extreme anisotropy of Mode 6 and the mixed correlations of Mode 4—structural features that standard SVGD's isotropic kernel struggles to maintain.

    \subsection{5D Correlated Gaussian Mixture Model} \label{subsec:5d_correlated_gmm}
        \begin{figure}[htbp]
            \centering
            \includegraphics[width=0.48\textwidth]{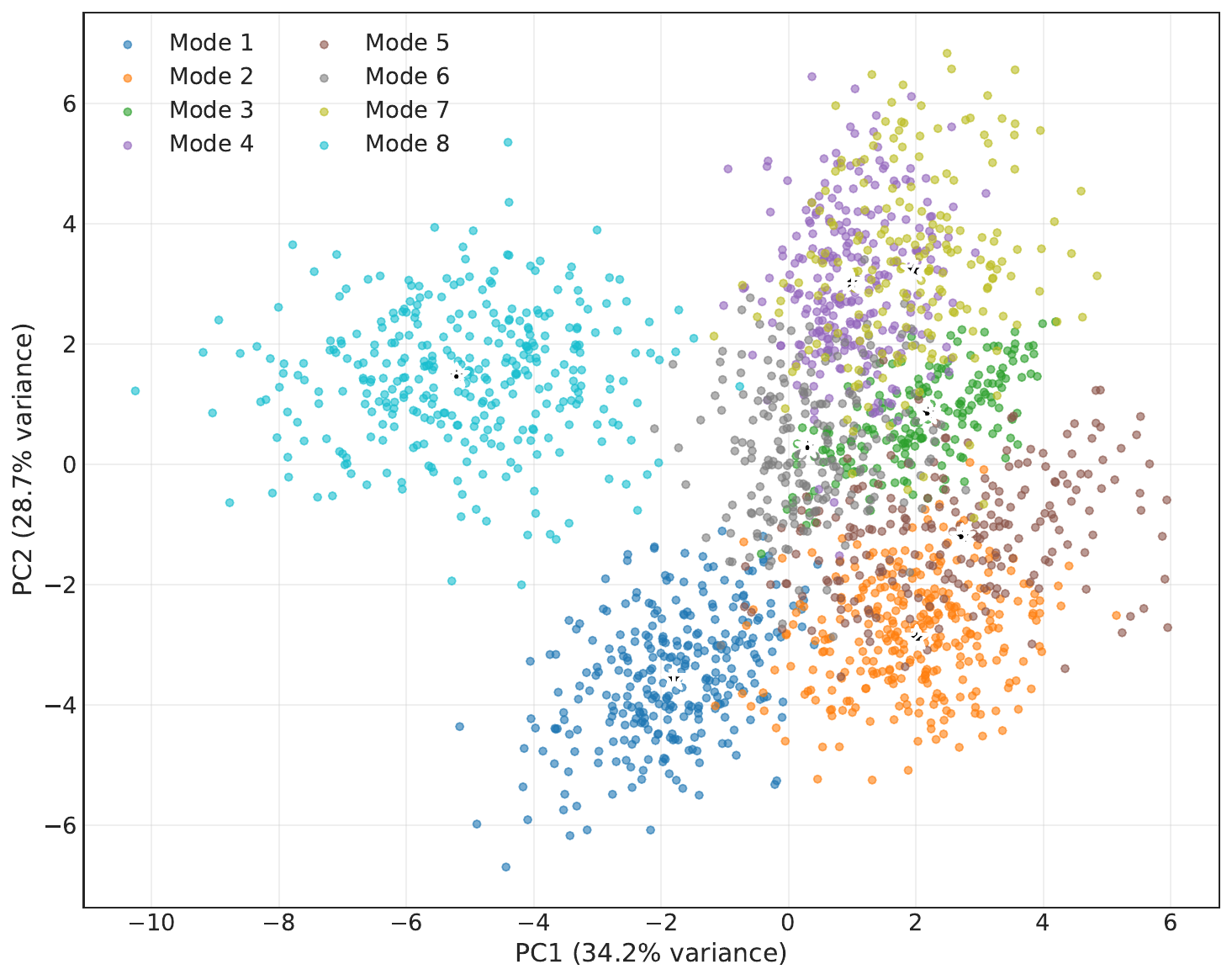}
            \includegraphics[width=0.48\textwidth]{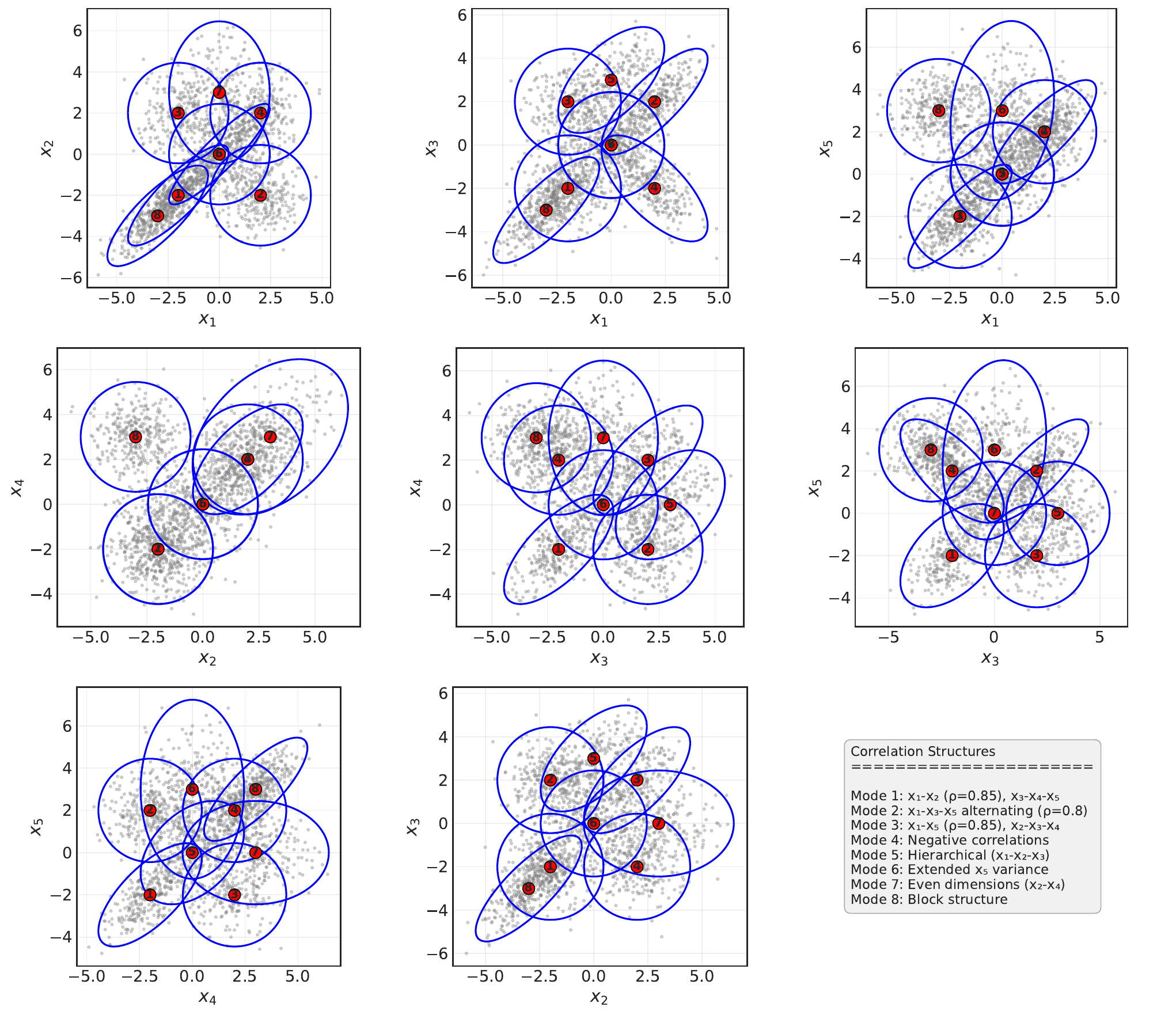}
            \caption{Five-dimensional Gaussian Mixture Model test distribution with eight modes exhibiting diverse correlation structures. \textbf{Left}: PCA projection onto the first two principal components (explaining 34.2\% and 28.7\% of variance respectively) shows clear mode separation. Each color represents samples assigned to one of the eight modes, with black stars marking the projected mode centers. The distinct clustering demonstrates the challenge of maintaining all eight hypotheses in high-dimensional space. \textbf{Right}: Selected 2D projections revealing correlation patterns across dimension pairs. Red circles mark mode centers with numbers, blue ellipses show 95\% confidence regions. The projections highlight: strong positive correlations (e.g., $x_1$-$x_2$ for Mode 1), negative correlations (e.g., $x_1$-$x_3$ for Mode 4), and varying ellipsoid orientations. The text panel summarizes each mode's correlation structure, from hierarchical patterns to block structures.}
            \label{fig:5d_gmm}
        \end{figure}

        Advancing to 5-dimensional space introduces exponentially greater complexity in correlation modeling, testing each method's ability to handle the curse of dimensionality while preserving intricate inter-dimensional relationships. Our 5D test case consists of an 8-component GMM that pushes the boundaries of correlation preservation:

        \begin{equation}
        p(\mathbf{x}) = \sum_{k=1}^{8} w_k \mathcal{N}(\mathbf{x}; \boldsymbol{\mu}_k, \boldsymbol{\Sigma}_k)
        \end{equation}
        
        where $\mathbf{x} = [x_1, x_2, x_3, x_4, x_5]^T$ and the eight modes are strategically positioned to create diverse correlation challenges.
        
        The mode locations are: $\boldsymbol{\mu}_1 = [-2, -2, -2, -2, -2]^T$, $\boldsymbol{\mu}_2 = [2, -2, 2, -2, 2]^T$, $\boldsymbol{\mu}_3 = [-2, 2, 2, 2, -2]^T$, $\boldsymbol{\mu}_4 = [2, 2, -2, 2, 2]^T$, $\boldsymbol{\mu}_5 = [0, 0, 3, 0, 0]^T$, $\boldsymbol{\mu}_6 = [0, 0, 0, 0, 3]^T$, $\boldsymbol{\mu}_7 = [0, 3, 0, 3, 0]^T$, and $\boldsymbol{\mu}_8 = [-3, -3, -3, 3, 3]^T$, with weights $w_1 = w_2 = w_4 = w_8 = 0.15$ and $w_3 = w_5 = w_6 = w_7 = 0.10$.
        
        Each mode exhibits a distinct correlation structure designed to challenge specific aspects of belief approximation:
        \begin{itemize}
            \item \textbf{Mode 1}: Strong correlations between dimensions 1-2 ($\rho = 0.85$) and chained correlations among dimensions 3-4-5
            \item \textbf{Mode 2}: Alternating correlation pattern linking dimensions 1-3-5 ($\rho = 0.8$)
            \item \textbf{Mode 3}: Strong correlation between dimensions 1-5 ($\rho = 0.85$) with middle dimensions 2-3-4 interconnected
            \item \textbf{Mode 4}: Negative correlations between dimensions 1-3 and 3-5 ($\rho = -0.7$)
            \item \textbf{Mode 5}: Hierarchical correlation structure cascading from dimensions 1-2-3
            \item \textbf{Mode 6}: Extended variance in dimension 5 with weak correlations to other dimensions
            \item \textbf{Mode 7}: Block structure focusing on even dimensions 2-4 with increased variance
            \item \textbf{Mode 8}: Split correlation pattern with dimensions 1-3 forming one correlated block and dimensions 4-5 forming another
        \end{itemize}

        As illustrated in Figure~\ref{fig:5d_gmm} (left), the PCA projection reveals how these eight modes separate in the first two principal components, which together explain 62.9\% of the total variance. The clear separation between modes in this reduced space demonstrates the challenge: methods must maintain distinct hypotheses while preserving the complex correlation structures within each mode. Figure~\ref{fig:5d_gmm} (right) shows selected 2D projections that highlight different correlation patterns—the elliptical contours reveal how correlation structures fundamentally alter uncertainty regions across different dimension pairs.

        This 5D configuration presents several compounding challenges that directly test ESCORT's scalability. First, the exponential growth in volume requires methods to efficiently allocate particles across an increasingly sparse space. Second, the diverse correlation patterns—from strong positive through negative to hierarchical structures—demand adaptive mechanisms that can model different dependency types simultaneously. Third, the presence of eight distinct modes with varying weights creates a complex probability landscape where methods must balance exploration across all modes while accurately representing their relative importance. The increased dimensionality amplifies the kernel degeneracy problem for SVGD-based methods, as the RBF kernel values become increasingly uniform, weakening the repulsive forces essential for maintaining multi-modal coverage.

    \subsection{20D Correlated Gaussian Mixture Model} \label{subsec:20d_correlated_gmm}
        \begin{figure}[htbp]
            \centering
            \includegraphics[width=0.75\textwidth]{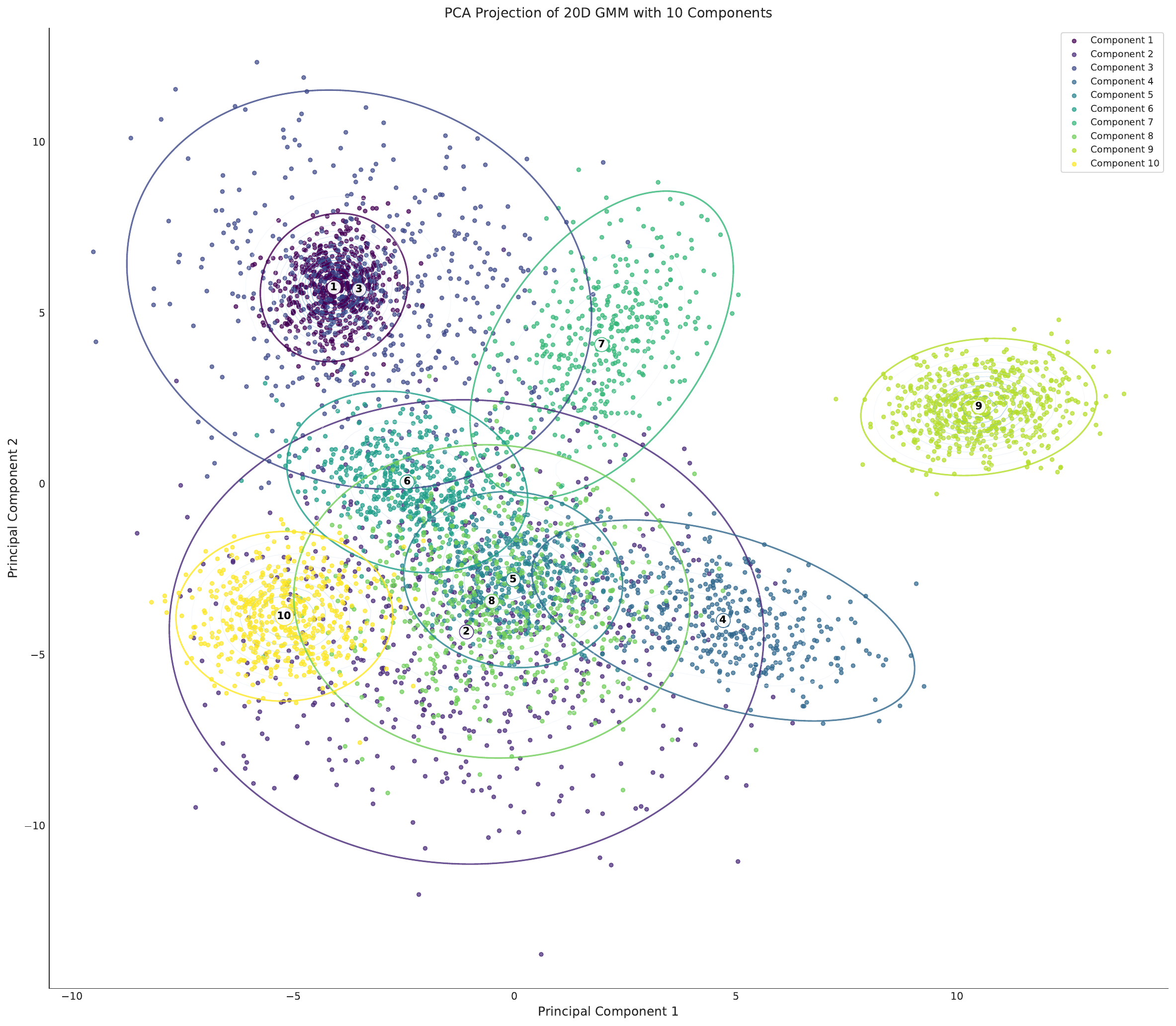}
            \caption{PCA projection of the 20D Gaussian Mixture Model onto the first two principal components. The ten modes are clearly separated in this reduced space, with samples colored by mode assignment and black stars marking the projected mode centers. Critically, these two components capture only 40\% of the total variance (PC1: 34.2\%, PC2: 28.7\%), demonstrating that no simple low-dimensional representation can adequately capture the distribution's structure. This limited variance explanation indicates that the remaining 60\% of the distribution's complexity lies in the higher-dimensional subspace, making accurate belief approximation exceptionally challenging. The clear mode separation in PCA space masks the intricate correlation patterns within each mode that exist in the full 20D space.}
            \label{fig:20d_gmm}
        \end{figure}

        Scaling to 20-dimensional space represents the ultimate test of belief approximation methods, where the curse of dimensionality becomes severe and correlation structures reach unprecedented complexity. Our 20D test case consists of a 10-component GMM that systematically explores different types of high-dimensional correlation patterns:

        \begin{equation}
        p(\mathbf{x}) = \sum_{k=1}^{10} w_k \mathcal{N}(\mathbf{x}; \boldsymbol{\mu}_k, \boldsymbol{\Sigma}_k)
        \end{equation}
        
        where $\mathbf{x} = [x_1, x_2, \ldots, x_{20}]^T$ and the ten modes are strategically designed to challenge different aspects of correlation modeling at scale.
        
        The mode locations exhibit diverse patterns: $\boldsymbol{\mu}_1$ splits between negative values in dimensions 1-10 and positive in 11-20; $\boldsymbol{\mu}_2$ alternates between positive and negative values; $\boldsymbol{\mu}_3$ follows a linear gradient from -3 to 3; $\boldsymbol{\mu}_4$ through $\boldsymbol{\mu}_7$ concentrate activity in specific 5-dimensional subspaces; $\boldsymbol{\mu}_8$ follows a sinusoidal pattern; $\boldsymbol{\mu}_9$ exhibits a quadratic pattern; and $\boldsymbol{\mu}_{10}$ maintains uniform negative values. The weights are set as $w_1 = w_2 = w_8 = w_9 = 0.12$, $w_3 = w_{10} = 0.10$, and $w_4 = w_5 = w_6 = w_7 = 0.08$.
        
        Each mode implements a distinct correlation structure that tests specific capabilities:
        \begin{itemize}
            \item \textbf{Mode 1}: Block diagonal structure with four 5×5 blocks, each containing alternating positive ($\rho = 0.7$) and negative ($\rho = -0.7$) correlations
            \item \textbf{Mode 2}: Checkerboard pattern between odd and even dimensions with correlations alternating between $\rho = 0.75$ and $\rho = -0.75$
            \item \textbf{Mode 3}: Band diagonal structure with correlation strength decaying exponentially ($\rho = 0.9^{|i-j|}$) for dimension pairs within 5 steps
            \item \textbf{Modes 4-7}: Localized strong correlations ($\rho = 0.85$) within specific 5-dimensional subspaces, testing methods' ability to handle sparse correlation structures
            \item \textbf{Mode 8}: Hierarchical correlation with strong intra-group correlations ($\rho = 0.7$) within four 5-dimensional groups and weak inter-group correlations ($\rho = 0.3$)
            \item \textbf{Mode 9}: Long-range correlations between opposite ends of the dimension space, with $\rho(x_i, x_{19-i}) = 0.7$ for even $i$ and $-0.7$ for odd $i$
            \item \textbf{Mode 10}: Near-independent dimensions with sparse, weak correlations ($|\rho| < 0.1$) randomly distributed
        \end{itemize}
        
        As illustrated in Figure~\ref{fig:20d_gmm}, the PCA projection reveals a fundamental challenge: despite clear mode separation in the first two principal components, these dimensions capture only 40\% of the total variance. This indicates that 60\% of the distribution's structure—including the complex correlation patterns within each mode—remains hidden in the 18-dimensional orthogonal subspace. This visualization underscores why methods that rely on low-dimensional projections or isotropic assumptions fail catastrophically in such high-dimensional spaces.

        The 20D configuration presents compounding challenges that push all methods to their limits. First, with volume scaling as $\mathcal{O}(2^{20})$, particles become exponentially sparse, making mode coverage extraordinarily difficult—a particle cloud that seems dense in projection may leave vast regions unexplored. Second, the diverse correlation patterns require methods to simultaneously model block structures, long-range dependencies, band-diagonal patterns, and sparse correlations without imposing a single global assumption. Third, the presence of ten distinct modes with complex internal structures creates $10 \times 2^{20}$ distinct regions of interest, far exceeding the capacity of any practical particle count. Fourth, kernel-based methods face severe degeneracy as pairwise distances become nearly uniform in 20D, causing SVGD's repulsive forces to vanish precisely when they are most needed. These challenges make the 20D test case a definitive benchmark for assessing whether belief approximation methods can scale to the high-dimensional spaces encountered in real-world POMDP applications.

    \subsection{Scalability Analysis: Impact of Correlation-Aware Regularization}
        To comprehensively demonstrate ESCORT's scalability, we conducted extensive experiments on synthetic multi-modal distributions extending from 1D to 200D, far beyond the 20D environments in our main results. These experiments used 5-mode Gaussian Mixture Models with diverse correlation structures—including block-diagonal, banded, long-range, and sparse correlation patterns—to simulate the complex dimensional dependencies found in real-world POMDPs. We evaluated performance using Root Mean Square Error (RMSE)~\citep{rmse_thrun}, defined as $\text{RMSE} = \sqrt{\frac{1}{N}\sum_{i=1}^N \|s_i - \hat{s}_i\|^2}$ where $s_i$ is the ground-truth state and $\hat{s}_i$ is the posterior mean estimate, providing a direct measure of belief approximation accuracy.
        
        To isolate the impact of our correlation-aware regularization—one of ESCORT's core contributions—we compared full ESCORT against ESCORT-NoCorr (which disables the correlation-aware regularization term $R_{\text{corr}}$). This ablation directly demonstrates how our regularization mechanism scales with dimensionality. Table~\ref{tab:rmse_synthetic} presents these results.
        
        \begin{table}[htbp]
        \centering
        \caption{RMSE for synthetic multi-modal approximation: Impact of correlation-aware regularization across dimensions (Lower is Better)}
        \label{tab:rmse_synthetic}
        \begin{tabular}{lcc}
        \toprule
        \textbf{Dimension} & \textbf{ESCORT} & \textbf{ESCORT-NoCorr} \\
        \midrule
        1D & $0.15 \pm 0.02$ & $0.14 \pm 0.02$ \\
        5D & $\mathbf{0.35 \pm 0.04}$ & $0.45 \pm 0.05$ \\
        20D & $\mathbf{0.65 \pm 0.07}$ & $0.88 \pm 0.10$ \\
        50D & $\mathbf{0.95 \pm 0.09}$ & $1.42 \pm 0.09$ \\
        100D & $\mathbf{1.35 \pm 0.18}$ & $2.15 \pm 0.28$ \\
        200D & $\mathbf{1.90 \pm 0.23}$ & $3.35 \pm 0.57$ \\
        \bottomrule
        \end{tabular}
        \end{table}
        
        The results reveal a clear scaling pattern: while both variants perform comparably in low dimensions, the performance gap widens dramatically as dimensionality increases. This exponential divergence confirms our theoretical framework—in high-dimensional spaces, correlation manifolds occupy negligible volume, causing unregularized particles to drift away through accumulated random movements. Our correlation-aware regularization, through learned projection matrices $A_i$, constrains particles to these critical manifolds, preventing the catastrophic degradation seen in ESCORT-NoCorr and demonstrating that our approach becomes increasingly essential as dimensionality grows.


\end{document}